\documentclass[10pt,twocolumn,letterpaper]{article}

\usepackage[pagenumbers]{iccv} 

\usepackage{array,multirow,graphicx}
\usepackage{makecell}
\usepackage{amssymb}
\usepackage{soul}
\usepackage{pifont}
\newcommand{\cmark}{\ding{51}}%
\newcommand{\xmark}{\ding{55}}%

\usepackage{float}
\newcommand{\STAB}[1]{\begin{tabular}{@{}c@{}}#1\end{tabular}}

\definecolor{cvprblue}{rgb}{0.21,0.49,0.74}
\definecolor{lightblue}{rgb}{0.9, 1, 1}
\usepackage{color, colortbl}
\usepackage[pagebackref,breaklinks,colorlinks,allcolors=cvprblue]{hyperref}

\def\paperID{5376} 
\def\confName{ICCV}
\def\confYear{2025}

\title{KeySync: A Robust Approach for Leakage-free \\ Lip Synchronization in High Resolution}

\author{Antoni Bigata$^{1}$ \qquad  \qquad Rodrigo Mira$^{1}$ \qquad Stella Bounareli \qquad Michał Stypułkowski$^{2}$ \\
Konstantinos Vougioukas$^{1}$ \qquad Stavros Petridis$^{1}$ \qquad Maja Pantic$^{1}$ \vspace{.2cm} \\
$^{1}$Imperial College London \qquad $^{2}$University of Wrocław \\
{\tt\small ab4522@imperial.ac.uk}
}

\begin{document}
\maketitle
\begin{abstract}
Lip synchronization, known as the task of aligning lip movements in an existing video with new input audio, is typically framed as a simpler variant of audio-driven facial animation. However, as well as suffering from the usual issues in talking head generation (\eg, temporal consistency), lip synchronization presents significant new challenges such as expression leakage from the input video and facial occlusions, which can severely impact real-world applications like automated dubbing, but are often neglected in existing works. To address these shortcomings, we present KeySync, a two-stage framework that succeeds in solving the issue of temporal consistency, while also incorporating solutions for leakage and occlusions using a carefully designed masking strategy. We show that KeySync achieves state-of-the-art results in lip reconstruction and cross-synchronization, improving visual quality and reducing expression leakage according to LipLeak, our novel leakage metric. Furthermore, we demonstrate the effectiveness of our new masking approach in handling occlusions and validate our architectural choices through several ablation studies. Code and model weights can be found at \href{https://antonibigata.github.io/KeySync/}{https://antonibigata.github.io/KeySync/}.
\end{abstract}    
\section{Introduction}
\label{sec:intro}

\begin{figure}
    \centering
    \includegraphics[width=1.0\linewidth]{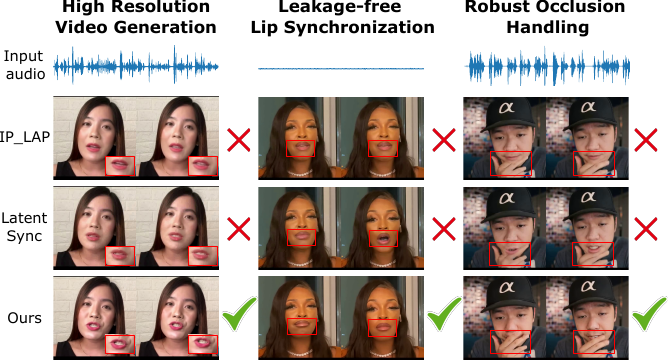}
    \caption{\textbf{KeySync's contributions.} Unlike existing methods, KeySync generates high-resolution lip-synced videos that are closely aligned with the driving audio while minimizing leakage from the input video and seamlessly handling facial occlusions.}
    \label{fig:intro_figure}
\end{figure}

Audio-driven facial animation has recently seen substantial progress with the introduction of new generative models such as Generative Adversarial Networks (GANs)~\cite{goodfellow2020generative, Vougioukas, zhou2019talking} and diffusion models~\cite{ho2020ddpm, stypulkowski2023, xu2024hallohierarchicalaudiodrivenvisual, wang2024V-Express, chen2024echomimiclifelikeaudiodrivenportrait}. In contrast, the adjacent field of lip synchronization (also known as lip-sync) has experienced comparatively slower advancements~\cite{guan2023stylesync, DINet, Prajwal}. This disparity is surprising given that lip-sync has similar applications, ranging from facilitating multilingual content production to enhancing virtual avatars~\cite{zhen2023human, zhan2023multimodal}. A potential reason for this slower progress is that while lip synchronization may seem like a simpler task than animating the full face from audio, it presents unique challenges that remain largely unaddressed.

One of the primary limitations of current methods is their low-resolution output, typically constrained to 256$\times$256, which has become a de facto standard. While this resolution may be computationally efficient, it significantly hinders real-world applicability, where higher resolution outputs are necessary for practical deployment. Another important limitation is that these methods struggle to maintain temporal consistency across generated frames. Most state-of-the-art approaches are frame-based, requiring additional mechanisms to enforce consistency, while other methods adopt a two-stage framework, where motion modelling is separated from pixel-level synthesis~\cite{maketalk, DiffDub, stylepres}. However, such decompositions do not inherently guarantee smooth transitions between frames, as each frame is generated independently, often leading to temporal discontinuities. 

Other approaches attempt to enforce temporal coherence using pre-trained perceptual models~\cite{latentsync} or an additional sequence discriminator~\cite{diff2lip}. Nevertheless, these methods offer only indirect control over frame-to-frame consistency, often resulting in subtle visual artifacts and unnatural mouth movements that degrade realism and limit practical usability. Finally, another proposed solution has been to condition generation on past frames~\cite{michaldub} to ensure that newly generated frames remain consistent with previous ones. However, this technique is prone to error accumulation over long sequences, further degrading temporal stability.

Beyond temporal consistency, a key but often overlooked issue is expression leakage, where models inadvertently retain facial expressions from the original input in the generated video. Regrettably, most existing works focus excessively on lip synchronization as a reconstruction task on paired audio-visual data, and neglect the cross-synchronization scenario, where a non-matching audio clip is used to re-animate the original video. As a consequence, they typically exhibit major expression leakage from the original video, severely degrading the synchronization between the generated video and the input audio in the latter scenario. Notably, this behaviour jeopardizes the viability of these models for applications such as automated dubbing, where the audio and video are naturally mismatched. 

To alleviate the issue of expression leakage, different masking strategies have been devised. Some methods mask only the mouth region while preserving facial areas such as the jaw and cheeks from the original videos, potentially leading to leakage since these regions also convey information about mouth movements~\cite{StyleLipSync, DINet}, while others adopt broader masks that risk discarding important contextual cues~\cite{zhang_musetalk_2024, VideoReTalking}. Remarkably, the impact of these masking strategies on generalization and robustness remains largely unexplored, and no consensus exists on the optimal approach. Lastly, another potential complication lies in occlusion handling. Most existing models assume an unobstructed view of the mouth, whereas, in the real world, occlusions caused by hands, objects, or motion blur are frequent. In practice, this means that the lack of explicit occlusion-handling mechanisms significantly limits the applicability of current models. 

To address these challenges, we propose KeySync, a two-stage lip synchronization framework that leverages recent advances in facial animation~\cite{keyface} to generate high-fidelity videos with lip movements that are temporally consistent and aligned with the input audio. To minimize leakage from the input video, we devise a masking strategy that adequately covers the lower face while retaining the necessary contextual regions. Furthermore, we augment this mask by excluding facial occlusions using a video segmentation model~\cite{sam2}, resulting in a method that can consistently handle occlusions without uncanny visual hallucinations. Our primary contributions, illustrated in Figure~\ref{fig:intro_figure}, can be summarized as:
\begin{itemize}
    \item \textbf{State-of-the-art lip synchronization}: KeySync achieves state-of-the-art lip synchronization performance at a resolution of $(512\times512)$,  surpassing the common $(256\times256)$ standard. It outperforms all competing methods in terms of quality and lip movement accuracy according to several objective metrics and a holistic user study. We observe particularly noticeable improvements in the cross-synchronization setting (where there is a mismatch between the input video and audio), enabling promising real-world applications such as automated dubbing.
    \item \textbf{A new strategy for occlusion handling}: We propose a new inference-time strategy for occlusion handling by excluding occluding objects from our mask automatically using a pre-trained video segmentation model. Through qualitative and quantitative analysis, we show that this method is consistently effective in handling occlusions.
    \item \textbf{A novel leakage metric}: To the best of our knowledge, we propose the first lip synchronization leakage metric (LipLeak), which computes the ratio of non-silent frames generated from a silent audio and a non-silent video, effectively measuring how often the lip movements from the input video leak into the generated video.
\end{itemize}

\section{Related Works}
\label{sec:formatting}

\paragraph{Audio-driven facial animation}
The goal of audio-driven facial animation methods is to generate high-quality talking head videos that preserve the identity of input faces while ensuring accurate lip movements synchronized with the input audio. Early GAN-based methods~\cite{DBLP:conf/bmvc/VougioukasPP18,Vougioukas, zhou2019talking, chung2017you} mostly focused on animating the speaker's facial expressions by introducing temporal constraints and expert discriminators to improve lip-sync accuracy. Later, several works~\cite{chen2020talking, zhang2022sadtalker, zhou2021pose} built on these approaches by incorporating head pose modelling to generate more realistic animations, but were prone to producing artifacts and unnatural motion.

Diffusion models~\cite{ho2020ddpm, rombach2022highresolutionimagesynthesislatent} have emerged as an alternative to GANs for audio-driven facial animation, demonstrating improved temporal consistency and video quality~\cite{du2023dae, xu2024vasa}. Several methods~\cite{stypulkowski2023, xu2024hallohierarchicalaudiodrivenvisual, wang2024V-Express, chen2024echomimiclifelikeaudiodrivenportrait} leverage video diffusion models~\cite{ho2022video, blattmann2023stablevideodiffusionscaling} for temporally consistent motion. Additionally, some works propose to condition the generation process on facial landmarks~\cite{wei2024aniportraitaudiodrivensynthesisphotorealistic} or 3D meshes \cite{zhang2023dream}. However, these approaches often produce non-realistic facial motion. Finally, a recent line of works~\cite{wei2024aniportraitaudiodrivensynthesisphotorealistic, wang2024V-Express, chen2024echomimiclifelikeaudiodrivenportrait} leverage ReferenceNet~\cite{animateanyone} to improve identity reconstruction, though at the cost of increased computational complexity. 

Recently, KeyFace~\cite{keyface} introduced a keyframe-based approach that predicts key poses and interpolates between them, enhancing identity preservation and temporal consistency. We follow their strategy, tailoring it to lip synchronization to ensure temporally consistent lip-sync animations that preserve the original identity without visible inpainting borders or expression leakage from the input video.

\begin{figure*}
  \centering
  \includegraphics[width=\textwidth]{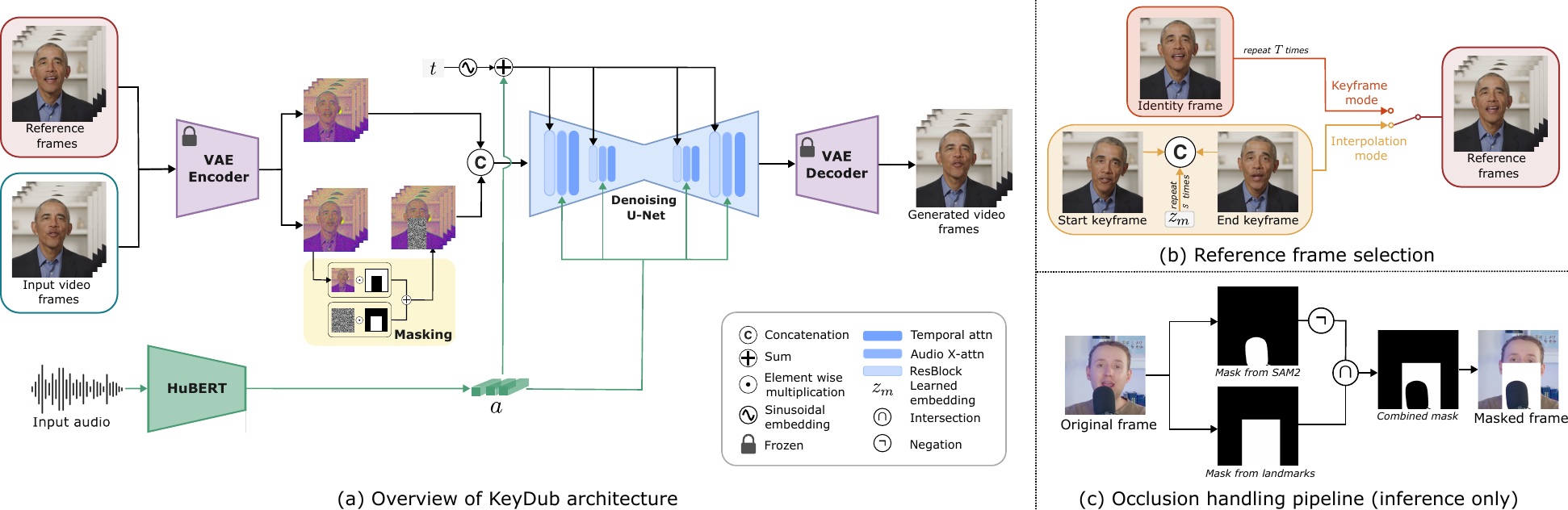}
    \caption{\textbf{Overview of the KeySync framework.} KeySync consists of two stages, both of which involve generating video using latent diffusion conditioned on an input video and audio, differing only in the reference frames selection, as described in (b). During keyframe generation, the model receives an identity frame  $x_{\text{id}}$, which is repeated and concatenated with the noised video input. During interpolation, the model is conditioned on two successive keyframes $z_{i}$ and $z_{i+1}$, along with intermediate learnable embeddings $z_m$. Both stages integrate audio embeddings $a$ from HuBERT~\cite{hubert}. In (c), we illustrate our occlusion handling pipeline, which we apply during inference.}
    \label{fig:architecture}
\end{figure*}

\paragraph{Audio-driven lip synchronization}
Lip synchronization methods focus on adjusting mouth movements to match the input audio while preserving other facial attributes, \ie, the head pose and upper face expressions. The first notable work, Wav2Lip \cite{Prajwal}, uses GANs to generate a sequence of frames from input video frames with the lower part of the face masked. To improve lip synchronization, Wav2Lip~\cite{Prajwal} leverages a pre-trained lip-sync expert model. In order to enhance realism and identity generalization, StyleSync~\cite{guan2023stylesync} and StyleLipSync~\cite{StyleLipSync} introduce StyleGAN2-based~\cite{karras2020analyzing} architectures, while DINet~\cite{DINet} performs spatial deformation on feature maps to improve visual quality. Finally, TalkLip~\cite{Zhong_2023_CVPR} proposes using contrastive learning based on a pre-trained lip-sync expert to enhance the quality and accuracy of generated lip region.

Recently, diffusion-based methods for lip synchronization have been introduced~\cite{diff2lip, DiffDub, michaldub}. Nevertheless, expression leakage from the input video, especially in cross-driving scenarios, remains an open issue. Several approaches attempt to mitigate this using different masking strategies~\cite{StyleLipSync, DINet, zhang_musetalk_2024, VideoReTalking}, but no consensus exists regarding the optimal masking method. Another challenge is temporal consistency, as many methods~\cite{maketalk, DiffDub, stylepres} operate on a frame-by-frame basis without explicit sequence modeling, leading to discontinuities. Some models are conditioned on past frames~\cite{michaldub}, but consequently suffer from cumulative error propagation, while others propose the use of pre-trained perceptual models~\cite{latentsync} or sequence discriminators~\cite{diff2lip} to enforce coherence, but are generally not sufficient. Finally, occlusion handling remains an open challenge, as most models assume clear visibility of the mouth, failing in real-world settings where occlusions from hands, objects, or motion blur occur.

Inspired by KeyFace~\cite{keyface}, we propose a two-stage lip synchronization framework ensuring robust, leakage-free cross-driving performance. Furthermore, our post-training occlusion-handling strategy further enhances robustness, rendering our model suitable for real-world applications.

\section{Method}
\label{sec:method}

In this section, we describe our proposed two-stage lip-sync approach, which builds upon KeyFace's facial animation framework~\cite{keyface}. Additionally, we discuss our masking strategy in Section~\ref{sec:masking} and present a new method for handling occlusions in Section~\ref{sec:occlusions}.

\subsection{Latent diffusion}

Diffusion models~\cite{ho2020ddpm, dhariwal2021diffusionmodelsbeatgans} progressively transform random noise into structured data by iteratively removing noise through a learned denoising process. Latent diffusion)~\cite{rombach2022highresolutionimagesynthesislatent} applies this denoising operation in a compressed, lower-dimensional latent space rather than in the high-dimensional pixel space, improving computational efficiency. Furthemore, the EDM framework~\cite{karras2022elucidatingdesignspacediffusionbased} defines the denoising operation of the denoiser $D_{\theta}$ as:
\begin{equation}
D_\theta(\mathbf{x}; \sigma) = c_{\text{skip}}(\sigma) \mathbf{x} + c_{\text{out}}(\sigma) F_\theta(c_{\text{in}}(\sigma) \mathbf{x}; c_{\text{noise}}(\sigma)),
\end{equation}
where $F_\theta$ denotes the trainable neural network and $\mathbf{x}$ represents the input. The terms $c_{\text{noise}}(\sigma)$, $c_{\text{out}}(\sigma)$, $c_{\text{skip}}(\sigma)$, and $c_{\text{in}}(\sigma)$ are scaling factors dependent on the noise level $\sigma$. These scaling factors dynamically adjust the magnitude and influence of noise at different stages of the denoising process, thereby improving the network's efficiency and robustness during diffusion.

\subsection{Leakage-proof masking} \label{sec:masking}

We frame the lip-sync task as a video inpainting problem~\cite{QuanCLYW24, SahariaCCLHSF022} in the latent space. The critical objective is to ensure the newly generated lip region does not reuse (or “leak”) cues from the original mouth shape that contradict the new audio. Specifically, we create a mask $M$ by computing facial landmarks~\cite{facealign} and isolating the lower facial region, extending slightly above the nose to cover any upper cheek movements that could otherwise convey information about lip movements, while still preserving overall facial identity. The mask also extends to the lower edge of the image, preventing any leakage from jaw movements. We find that this mask strikes an appropriate balance between the two types of masks presented in prior works, namely:
\begin{itemize}
    \item \textbf{Full lower-face masks}~\cite{stylepres, DiffTalk, diff2lip, SyncTalkFace}, which can obscure too much context, risking issues with identity and natural facial continuity;
    \item \textbf{Mouth-only masks}~\cite{DINet, DiffDub, StyleLipSync, VideoReTalking}, which can inadvertently leak lower face expressions because residual mouth movements or shading remain visible to the model.
\end{itemize}

\subsection{Two-stage video generation}
\label{sec:pipeline}
Our approach is illustrated in Figure~\ref{fig:architecture}. We follow KeyFace's two-stage procedure~\cite{keyface} to generate lip-synced animations from a masked input video and a driving audio clip. We feed the video frames $\{x_t\}_{t=1}^T$ and their corresponding noised versions $\{x^n_t\}_{t=1}^T$, into our VAE encoder~\cite{blattmann2023stablevideodiffusionscaling} $\mathcal{V}$ to obtain latent representations $\{z_t\}_{t=1}^T$ and $\{z^n_t\}_{t=1}^T$, respectively. Then, using the predefined mask M, we define the input to the U-Net as:
\begin{equation}
    z_t^m = M \odot z^n_t + (1 - M) \odot z_t,
\end{equation}
where $\odot$ denotes element-wise multiplication.
Given the corresponding audio segments $\{a_t\}_{t=1}^T$, our goal is to generate a new set of frames $\{\hat{x}_t\}_{t=1}^T$ in which the lip movements are fully synchronized with the audio.
Unlike previous approaches that either generate all frames end-to-end~\cite{StyleLipSync, talklip, latentsync} or explicitly disentangle motion and appearance~\cite{DiffDub, stylepres, maketalk}, we propose a two-stage strategy for lip synchronization.

\paragraph{Keyframe Generation}
We first generate a sparse set of keyframes capturing the essential lip movements for the entire audio sequence. These serve as anchor points, ensuring that each keyframe accurately reflects the phonetic content of the audio while preserving the user’s identity (through an identity frame $\mathbf{x}_{id}$). We build on Stable Video Diffusion (SVD)~\cite{blattmann2023stablevideodiffusionscaling}, a latent diffusion model that operates on batches of frames with a 2D U-Net and additional 1D temporal layers. Specifically, we produce $T$ keyframes, each spaced $S$ frames apart,
$\{\hat{x}_{t_k}\}_{k=1}^T$, where $t_k = k \cdot S$.

\paragraph{Interpolation}
Next, we interpolate between successive keyframes to achieve smooth, temporally coherent motion. Using the same diffusion backbone, we condition on keyframe pairs ($\hat{z}{t_i}$, $\hat{z}{t_{i+1}}$) to generate the intermediate frames. Specifically, we construct the sequence:
\begin{equation}
    s = \{z_{t_i}, \underbrace{z_m, \dots, z_m}_{\text{repeat} \ S \ \text{times}}, z_{t_{i+1}} \},
\end{equation}
where $z_m$ is a learnable embedding that represents the missing frames. This approach enables us to model the temporal dynamics of the video directly, without requiring additional synchronization losses~\cite{diff2lip}, temporal perceptual models~\cite{latentsync}, or motion-specific frames~\cite{michaldub}. \\

Throughout both steps, we rely on the audio encoder  $A$, HuBERT~\cite{hubert}, to transform raw audio into a learned representation. This audio embedding is integrated into the model via audio attention blocks, where the embeddings are fed into the U-Net's cross-attention layers, and also via the timestep embeddings, where the audio features are passed through an MLP and added to the diffusion timestep embeddings $\mathbf{t}_s \in \mathbb{R}^{C_s}$, resulting in $ \mathbf{t}^{\prime}_s = \mathbf{t}_s + \text{MLP}(a)$. This dual mechanism enhances alignment between video and audio frames, leading to improved lip synchronization.

\subsection{Losses}

We adopt the loss formulation from \cite{karras2022elucidatingdesignspacediffusionbased}, defined as:
\begin{equation}
    \mathcal{L}_{latent} = \mathbb{E}_{x, c, t, \sigma} \left[w_t \left\|F_{\theta}(z^m_t; c, \sigma_t) - z_t\right\|_2^2\right],
\end{equation}
where $w_t$ is a predefined weighting function, $F_{\theta}$ represents the model to be trained, $\sigma_t$ denotes the noise level, and $c$ the conditioning inputs to the model.
We find that this loss alone is sufficient to achieve good lip synchronization and high-quality video generation. However, working solely in the compressed latent space can make it difficult for the model to retain fine semantic details \cite{show1}, which are critical for real-world lip synchronization tasks where preserving the nuances of the mouth region is essential. To address this, we introduce an additional $L_2$  loss in the RGB space. This requires decoding the generated latent sequence using the VAE decoder $\mathcal{V}$, resulting in:
\begin{equation}
    \mathcal{L}_{rgb} = \mathbb{E}_{x, c, t, \sigma} \left[w_t \left\|\mathcal{V}(F_{\theta}(z^m_t; c, \sigma_t)) - x_t\right\|_2^2\right].
\end{equation}
To optimize memory efficiency, we apply $\mathcal{L}_{rgb}$ to a randomly selected frame from the sequence, which we found to be sufficient for maintaining perceptual quality.
The final loss function is then:
\begin{equation}
    \mathcal{L}_{total} = M \cdot \lambda(t)(\mathcal{L}_{latent}(\hat{z},z) + \lambda_2 \mathcal{L}_{rgb}(\hat{x},x)),
\end{equation} 
where $\lambda(t)$ is a weighting factor dependent on the diffusion timestep  $t$, as defined in \cite{karras2022elucidatingdesignspacediffusionbased}. Importantly, we ensure that only the generated region contributes to the loss computation by masking the region of interest.

\subsection{Handling occlusions}
\label{sec:occlusions}

Occlusions are a critical yet often overlooked challenge in lip synchronization. Even advanced models can produce unnatural results if occlusions in the original video, such as a hand or microphone covering the mouth, are not properly accounted for. A common issue arises when an occlusion overlaps with the mouth region during masking, often causing the model to incorrectly generate the mouth over the occluding object, resulting in unnatural boundary artifacts.

To address this, we propose an inference-time solution capable of handling any type of occlusion without additional training. Explicitly training a model for occlusion handling is impractical due to the vast range of possible occlusions and their inherent misalignment with speech, making them hard for the model to learn. Instead, we introduce a preprocessing pipeline that first segments the occluding object using a state-of-the-art zero-shot video segmentation model~\cite{sam2}, generating a mask  $M_{obj}$  of the occlusion. We then refine the original mask  M  by excluding the occlusion:
\begin{equation}
M = M \cap \neg M_{obj},
\end{equation} where $\cap$ denotes intersection and $\neg$ denotes logical negation. Since our model supports free-form masks, as in \cite{repaint}, it can seamlessly reconstruct the mouth region while preserving the occluding object, ensuring visually coherence.

\section{Experiments}
\label{sec:experiments}

\subsection{Datasets}

We train our model on three widely used audiovisual datasets: HDTF~\cite{hdtf}, CelebV-HQ~\cite{zhu2022celebvhq}, and CelebV-Text~\cite{yu2022celebvtext}. While HDTF is a carefully curated dataset, CelebV-Text and CelebV-HQ prioritize quantity and include many low-quality videos, instances where the speaker is out of frame, and abrupt cuts. To address these issues, we implement a curation and preprocessing pipeline that we use to refine the contents of CelebV-Text and CelebV-HQ, and then combined them with HDTF for training. This pipeline is described in detail in the supplementary material.

For evaluation, we focus on the cross-sync task, the primary use case for lip-sync models, where the input audio comes from a different video than the one being generated. We randomly select 100 test videos from CelebV-Text, CelebV-HQ, and HDTF and swap their audio tracks. Additionally, to ensure consistency with prior works, we also report reconstruction results for the same 100 videos.

\subsection{Evaluation metrics}
\label{sec:metrics}

For evaluation, we rely on a set of no-reference metrics, as we found they correlate better with human perception, particularly in the cross-sync task. For image quality, we use CMMD~\cite{cmmd}, an improved version of FID, along with a version of TOPIQ~\cite{topiq} trained on a facial dataset from~\cite{pyiqa}. However, we found that these methods do not effectively capture image blurriness. To address this, we incorporate a common blurriness evaluation method: the variance of Laplacian (VL)~\cite{laplacian}. For video quality and temporal consistency, we use FVD~\cite{fvd}. Additionally, we introduce LipLeak, a new metric for leakage detection, and rely on LipScore~\cite{keyface} to evaluate lip synchronization, which has been shown to be more effective than the offset and confidence scores produced by the commonly used SyncNet model~\cite{Prajwal}. Further details are provided in the supplementary material.

\paragraph{LipLeak} While leakage is a known issue in lip-sync models, there is currently no direct method to quantify it. We propose a simple yet effective evaluation approach based on feeding silent audio and non-silent video into the model. Given the silent audio, it can be assumed that any frame where the mouth is open is the result of expression leakage from the non-silent input video. To this end, we assess leakage by computing the proportion of time the mouth is open using the mouth aspect ratio (MAR)~\cite{mouthopen}. MAR is defined as the ratio between the vertical distance between the upper and lower lip landmarks and the horizontal distance between the corner lip landmarks. By applying an empirically determined threshold (set to 0.25) to distinguish open-mouth states, we obtain a quantitative measure of lip movement leakage. This ratio provides a direct measure of a model’s susceptibility to lip movement leakage, hence we denote it as LipLeak.

\subsection{User study}

While the metrics above offer an objective evaluation, they do not always align with human perception. To address this, we conduct a user study where participants compare randomly selected video pairs based on lip synchronization, temporal coherence, and visual quality. We then rank the performance of each model using the Elo rating system~\cite{elo1978rating}, and apply bootstrapping~\cite{chiang2024chatbot} for robustness. Further details are provided in the supplementary material.

\section{Results}
\label{sec:results}

\begin{table*}[ht]
\centering
\begin{tabular}{llccccccc}
\toprule
                                          & \multirow{2}{*}{Method}      & \multirow{2}{*}{\makecell{CMMD $\downarrow$}} & \multirow{2}{*}{\makecell{TOPIQ $\uparrow$}} & \multirow{2}{*}{\makecell{VL $\uparrow$}} & \multirow{2}{*}{\makecell{FVD $\downarrow$}} & \multirow{2}{*}{\makecell{LipScore $\uparrow$}} 
                                          & \multirow{2}{*}{\makecell{LipLeak $\downarrow$}} & \multirow{2}{*}{\makecell{Elo $\uparrow$}} \\ \\ \midrule
\multirow{6}{*}{\STAB{\rotatebox[origin=c]{90}{Reconstruction}}}    & DiffDub~\cite{DiffDub}     &  0.403  &       0.44    & 37.12 &429.07   &   0.34    & -  &   1014  \\         
& IP\_ LAP~\cite{Zhong_2023_CVPR}     & \underline{0.091}   &  \underline{0.49}   &  37.77 &\underline{282.02}    &  0.36    &   -    &  1007     \\
& Diff2Lip~\cite{diff2lip}   &   0.225     &   0.48    & 35.84 & 555.08   &   0.49    &  -   &  886    \\
                                          & TalkLip~\cite{talklip}   & 0.230   &  0.39     & 29.07  &608.92   & \textbf{0.58}   &    -   & 920       \\ 
                                          & LatentSync~\cite{latentsync}   & 0.319   &  0.41     & \underline{45.23}  &343.90   & \underline{0.52}   &    -   &    \underline{1052}    \\ 
                                          & \cellcolor{cvprblue!40}KeySync &  \cellcolor{cvprblue!40}\textbf{0.064} &  \cellcolor{cvprblue!40}\textbf{0.58}  &  \cellcolor{cvprblue!40}\textbf{70.32} & \cellcolor{cvprblue!40}\textbf{191.21}  &  \cellcolor{cvprblue!40}0.46    & \cellcolor{cvprblue!40}- & \cellcolor{cvprblue!40}\textbf{1120} \\ \hline
\multicolumn{1}{c}{\multirow{6}{*}{\STAB{\rotatebox[origin=c]{90}{Cross-sync}}}} 

& DiffDub~\cite{DiffDub}         &  0.408  &    0.44   & 37.05 &420.66     & \underline{0.34}    & 0.56     &  947       \\
\multicolumn{1}{c}{}                      & IP\_\ LAP~\cite{Zhong_2023_CVPR}    &  \underline{0.093}  &   0.49  & 35.32 &\underline{294.66}   &  0.17    &  0.28     &     1031      \\
\multicolumn{1}{c}{}                      & Diff2Lip~\cite{diff2lip}   & 0.231  & \underline{0.48}   &33.97 &601.68  &  0.16    & \underline{0.25}    &       878        \\

\multicolumn{1}{c}{}                      & TalkLip~\cite{talklip}    & 0.201  &  0.42  & 24.80  &704.93   &  0.30  &  0.66      &    911    \\
& LatentSync~\cite{latentsync}   & 0.325  &  0.41     & \underline{45.95}  &361.57   & 0.14   &    0.33   &   \underline{1086}    \\ 
\multicolumn{1}{c}{}                      & \cellcolor{cvprblue!40}KeySync &  \cellcolor{cvprblue!40}\textbf{0.070}
 &  \cellcolor{cvprblue!40}\textbf{0.58}  &  \cellcolor{cvprblue!40}\textbf{73.04} & \cellcolor{cvprblue!40}\textbf{206.32}  &  \cellcolor{cvprblue!40}\textbf{0.48}    & \cellcolor{cvprblue!40}\textbf{0.16}   & \cellcolor{cvprblue!40}\textbf{1145}  \\ 
                                         \bottomrule
\end{tabular}

\caption{\textbf{Quantitative comparison with other works} on reconstruction and cross-synchronization performance. The best results are highlighted in \textbf{bold}, while the second-best results are \underline{underlined}. All metrics are described in Section~\ref{sec:metrics}.
}
\label{tab:metrics_comparison}
\end{table*}

In this section, we present a comprehensive evaluation of our model’s performance against baselines, along with several ablations to assess the impact of key components.

\subsection{Comparison with other works} \label{sec:quantitative}
\paragraph{Quantitative analysis.}

We evaluate our method alongside five competing approaches in Table~\ref{tab:metrics_comparison}. The evaluation is conducted in two settings: reconstruction, where videos are generated using the same audio as in the original video, and cross-sync, where the audio is taken from a different video. The latter is particularly relevant as it better reflects real-world applications such as automated dubbing, where the driving audio is typically not aligned with the input video. 

KeySync consistently outperforms other methods in visual quality and temporal consistency for both reconstruction and cross-sync tasks, as reflected in the higher VL scores and lower FVD. Regarding lip synchronization, while most methods experience a drop in LipScore during cross-sync, our approach maintains consistent performance, as evidenced by similar LipScores in both reconstruction and cross-sync. In the reconstruction setting, some methods achieve a higher LipScore, but this is primarily due to expression leakage. This hypothesis is also supported by the high LipLeak scores exhibited by these models. An interesting observation arises with DiffDub: in the cross-sync setting, the model generates random lip movements, which LipScore mistakenly interprets as improved synchronization, but LipLeak exposes as leakage. Our method is also preferred by human evaluators in both settings, as shown by the higher Elo rankings, highlighting the effectiveness of our approach according to human perception.

\paragraph{Qualitative analysis.}

\begin{figure}
    \centering
    \includegraphics[width=1.0\linewidth]{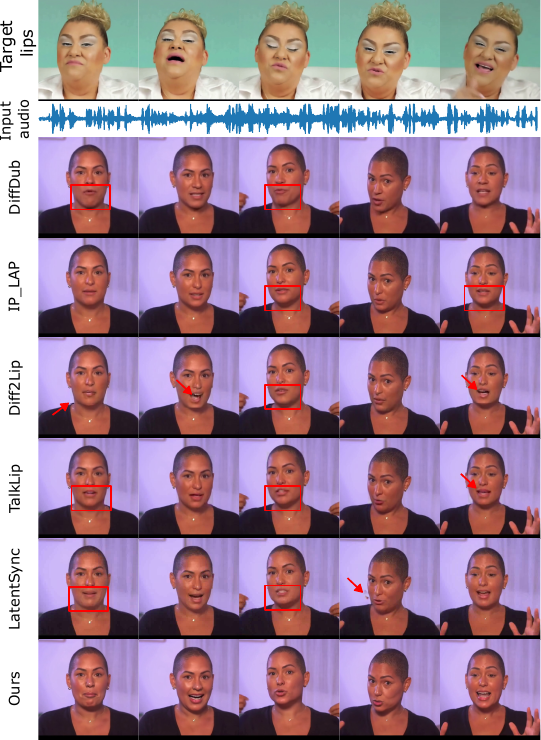}
    \caption{\textbf{Qualitative comparison with other works.} The top row (“Target lips”) shows lip movements corresponding directly to the provided audio input, and can therefore be seen as the target for the lips in the generated videos.}
    \label{fig:qualitative}
\end{figure}

Figure~\ref{fig:qualitative} presents results from all models for a cross-sync example. We see that KeySync more accurately follows the lip movements corresponding to the input audio. While LatentSync and Diff2Lip also appear to align somewhat with the target lip movements, they fail to generate certain vocalizations correctly and exhibit visual artifacts (highlighted on the figure via red squares and arrows, respectively), limiting their practical usability. Additionally, we notice that most methods produce minimal lip movement or fail to open the mouth sufficiently. This can be attributed to expression leakage, where the model receives conflicting signals from the original video and the new audio, making it difficult to generate a coherent and natural-looking mouth region.

\paragraph{Leakage.}

\begin{figure}
    \centering
    \includegraphics[width=1.0\linewidth]{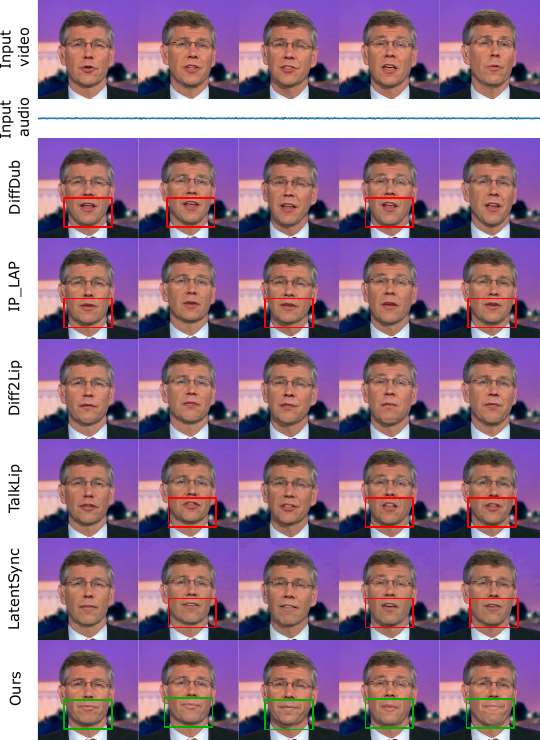}
    \caption{\textbf{Qualitative leakage comparison.} We condition the models on silent audio and non-silent video (first row).}
    \label{fig:leakage}
\end{figure}

\begin{figure}
    \centering
    \includegraphics[width=\linewidth]{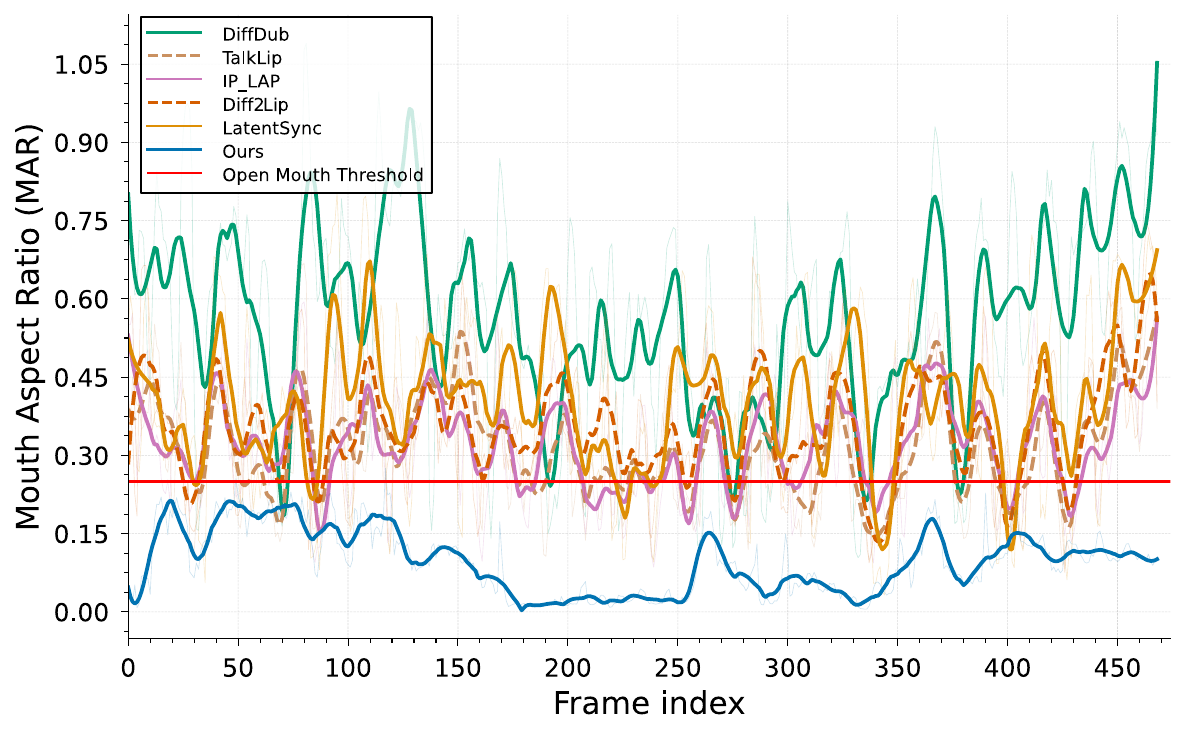}
    \caption{\textbf{MAR over time.} If MAR exceeds the threshold, the mouth is considered open, indicating leakage.}
    \label{fig:leakage_plot}
\end{figure}

As discussed in Section~\ref{sec:metrics}, we compute LipLeak by generating a video using a silent audio input. Since the audio contains no speech, the mouth should remain closed throughout the entire duration. However, in practice, we observe that this is not always the case, as the non-silent expressions in the input video can leak into the generated video. Figure~\ref{fig:leakage} presents examples of such generated videos alongside the original input video and silent audio. We observe that all methods, except ours and Diff2Lip, exhibit several frames where the mouth is open (highlighted by red squares around the mouths) due to expression leakage from the input video. While Diff2Lip manages to keep the mouth closed, we notice significant blending artifacts across all frames, highlighting the model’s struggle to generate silent frames when the original video contains speech. Additionally, in Figure~\ref{fig:leakage_plot}, we visualize the mouth aspect ratio (MAR) of all methods over time. Since MAR is a ratio, it remains independent of scale, ensuring a fair comparison across different methods. The results show that baseline methods frequently exceed the MAR threshold for an open mouth, indicating persistent leakage. In contrast, KeySync consistently maintains a MAR below the threshold, demonstrating its effectiveness in preventing unwanted lip movements when no speech is present.

\paragraph{Occlusion handling.}
\begin{figure*}
    \centering
    \includegraphics[width=\textwidth]{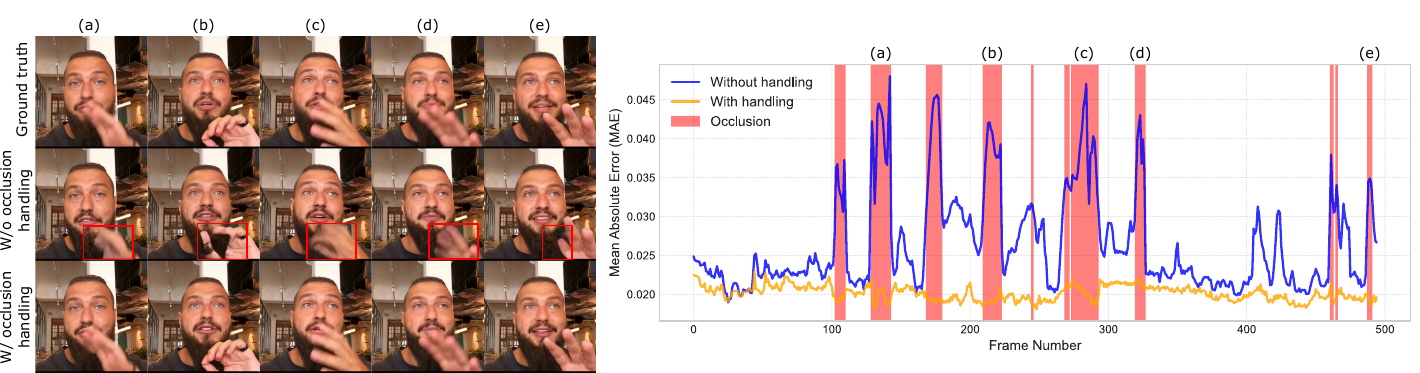}
    \caption{\textbf{Occlusion handling comparison.} We present qualitative results on the left and quantitative results on the right.}
    \label{fig:occlusions}
\end{figure*}

We propose a technique to handle occlusions at inference time, as defined in Section~\ref{sec:occlusions} and illustrated in Figure~\ref{fig:architecture}. The effectiveness of our approach is demonstrated in Figure~\ref{fig:occlusions}. On the left side, we observe that without proper occlusion handling, the model fails to reconstruct the occluded regions accurately, leading to unwanted artifacts, particularly around the hand. In contrast, with our proposed method, the hand is correctly represented while maintaining lip synchronization. This improvement is further highlighted in the right part of the figure, where we visualize the mean absolute error between the generated video and the ground-truth. Without occlusion handling, we observe noticeable error spikes during occlusions. 
Additional details are provided in the supplementary material.

\subsection{Ablation studies}


\paragraph{Architecture.}

We evaluate two alternate versions of our pipeline and present the results in Table~\ref{tab:architecture_ablation}. The first approach replaces our two-stage pipeline with a single-stage model, where the network directly generates a sequence of frames, and a longer video is produced by concatenating multiple sequences with a one-frame overlap to ensure temporal consistency. The second approach keeps the two-stage logic,  but generates keyframes using an image-based model rather than producing them all simultaneously with temporal layers. We find that the one-stage model achieves reasonable visual quality according to CMMD, but sees a sharp decline in FVD and LipScore, highlighting the importance of our keyframe interpolation technique for generating smooth, well-synchronized lip movements. Similarly, without the temporal layers, the interpolation model struggles to maintain coherence, leading to significant degradation across all three metrics.

\paragraph{Audio encoder.}
\begin{table}[t]
\centering
\resizebox{\columnwidth}{!}{
\begin{tabular}{ccccc}
\toprule
Two-stage & Temp. layers & CMMD $\downarrow$& FVD $\downarrow$ & LipScore $\uparrow$ \\ \midrule
  \xmark & \cmark      &    \underline{0.085}              & \underline{395.45} & 0.32   \\
  \cmark     & \xmark  & 0.142             & 618.27 & \underline{0.39} \\
   \rowcolor{Gray!40} \cmark     & \cmark  & \textbf{0.070}                  &  \textbf{206.32} &  \textbf{0.48} \\
\bottomrule
\end{tabular}}
\caption{\textbf{Architecture ablation} in the cross-sync setting.}
\label{tab:architecture_ablation}
\end{table}
\begin{table}[t]
\centering
\begin{tabular}{lcccc}
\toprule
Audio backbone     & FVD $\downarrow$ & LipScore $\uparrow$ & LipLeak $\downarrow$ \\ \midrule
Whisper~\cite{whisper}    &  207.41  &  0.47   &  0.18  \\
Wav2vec2~\cite{wav2vec2}         & \textbf{201.13}    & 0.45  & 0.19  \\ 
 WavLM~\cite{wavlm}    &  218.08   & \textbf{0.48}  & 0.17  \\
\rowcolor{Gray!40} HuBERT~\cite{hubert}        & \underline{206.32}    & \textbf{0.48}  & \textbf{0.16}  \\
\bottomrule
\end{tabular}
\caption{\textbf{Audio encoder ablation} in the cross-sync setting.}
\label{tab:abl_audio}
\end{table}
\begin{table}[t]
\centering
\resizebox{\columnwidth}{!}{%
\begin{tabular}{lcccc}
\toprule
Mask     &        CMMD $\downarrow$& FVD $\downarrow$& LipScore $\uparrow$ & LipLeak $\downarrow$ \\ \midrule
Mouth-only      &  0.077  &  \underline{200.71}   & 0.23  &  0.38 \\
Full lower-face   &  0.743   &  219.96   & 0.35  & \underline{0.28}  \\
Ours (nose-level)   &  \underline{0.071}   &  \textbf{199.39}   &  \underline{0.34} & 0.35 \\
\rowcolor{Gray!40} Ours &  \textbf{0.070}   &  206.32   & \textbf{0.48}  & \textbf{0.16}  \\
\bottomrule
\end{tabular}}
\caption{\textbf{Mask ablation} in the cross-sync setting.}
\label{tab:abl_mask}
\end{table}

We also investigate the impact of different audio encoders on the generated videos, as shown in Table~\ref{tab:abl_audio}. We see that Wav2vec2~\cite{wav2vec2} produces marginally higher video quality, as indicated by its lower FVD score. However, this comes at the expense of lip synchronization, as reflected in its lower LipScore. With WavLM~\cite{wavlm}, we achieve a LipScore comparable to HuBERT~\cite{hubert}, but at the cost of worse video quality. In contrast, HuBERT not only maintains a strong LipScore but also achieves the lowest LipLeak, indicating its effectiveness in mitigating expression leakage. Based on these findings, we select HuBERT as our default audio encoder.

\paragraph{Mask.}
\begin{figure}
    \centering
    \includegraphics[width=\linewidth]{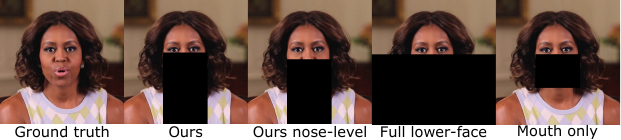}
    \caption{\textbf{Examples of different masking techniques.}}
    \label{fig:masking}
\end{figure}

Finally, we investigate the impact of different masking techniques (illustrated in Figure~\ref{fig:masking}) in Table~\ref{tab:abl_mask}. Using a mouth-only mask yields better video quality since it minimizes facial obstruction. However, this approach suffers from severe leakage, as indicated by its low LipScore and high LipLeak. This occurs because the model can infer mouth movements from the mask’s position over time rather than learning proper synchronization with the driving audio. Conversely, masking the entire lower face effectively reduces leakage, but severely harms image and video quality, as the model is forced to reconstruct unrelated background elements. Our proposed box-style masking offers a balanced trade-off between these two extremes, achieving the best overall performance. Additionally, we demonstrate that extending the mask up to the eye region is crucial in maximizing LipScore, as the cheeks convey important cues about mouth movements and can therefore cause leakage. 
\section{Conclusion} \label{sec:conclusion}
In this paper, we propose KeySync, a state-of-the-art lip synchronization approach based on a two-stage video diffusion model. We show that, unlike other methods, KeySync generates high-resolution videos which are temporally coherent and closely aligned with the driving audios. Furthermore, by applying a new masking strategy, we show that our model successfully minimizes expression leakage from the input video, while also being robust to facial occlusions that may occur in the wild. We hope that these improvements will enable the use of lip synchronization models in applications such as automated dubbing, which can help eliminate language barriers at scale.

{
    \small
    \bibliographystyle{ieeenat_fullname}
    \bibliography{main}
}
\clearpage
\setcounter{page}{1}
\maketitlesupplementary
\appendix

\section{Datasets}

\subsection{Curation and preprocessing}

\begin{figure}
    \centering
    \includegraphics[width=\linewidth]{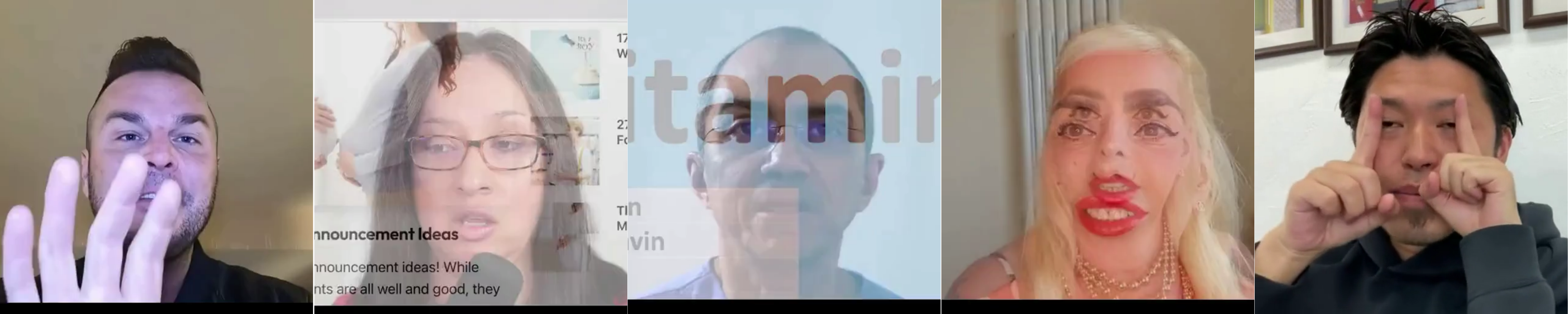}
    \caption{\textbf{Examples of problematic videos in CelebV-HQ and CelebV-Text.}}
    \label{fig:bad_examples}
\end{figure}

When working with in-the-wild datasets such as CelebV-HQ~\cite{zhu2022celebvhq} and CelebV-Text~\cite{yu2022celebvtext}, we observed that a significant portion of the data is of suboptimal quality. Common issues include visible hands, camera movement, editing artifacts, and occlusions. Additionally, some samples exhibit lower resolution than advertised. Examples of these issues are illustrated in Figure~\ref{fig:bad_examples}. During training, we found that such videos negatively impacted model performance because their visual content correlates poorly with the corresponding audio. To address these challenges, we developed a data curation pipeline comprising the following steps:

\begin{itemize}
\item Extract videos at 25 FPS and single-channel audio at 16 kHz.
\item Discard low-quality videos based on HyperIQA~\cite{Su_2020_CVPR_hyperiqa} scores below 0.4. Each video’s score is computed as the average of nine evaluations: selecting the first, middle, and last frames, each evaluated on three random crops.
\item Detect and segment scenes using \href{https://github.com/Breakthrough/PySceneDetect}{PySceneDetect}.
\item Remove clips without active speakers using Light-ASD~\cite{Liao_2023_CVPR_lightSDA} indicated by the score below 0.75.
\end{itemize}

\subsection{Data statistics}

Table~\ref{tab:dataset} describes the training/evaluation data used in this paper, specifying the number of speakers, videos, average video duration, and total duration for each dataset. Additionally, to illustrate the impact of our data curation pipeline, we present Table~\ref{tab:uncleaned_dataset}, which details the statistics of the datasets before curation. Overall, we discard roughly 75\,\% of the original videos. Please note that CelebV-HQ and CelebV-Text videos were split into shorter chunks during pre-processing, hence the higher video count in Table~\ref{tab:dataset}.

\begin{table}[ht]
\centering
\resizebox{\columnwidth}{!}{
\begin{tabular}{lrrrr}
\toprule
Dataset         & \# Speakers & \# Videos & \multicolumn{2}{c}{Duration} \\ \cmidrule(lr){4-5}
                &             &           & Avg. (sec.) & Total (hrs.)  \\ \midrule
HDTF~\cite{hdtf}       &   264       &   318    &    139.08      &       12     \\
CelebV-HQ~\cite{zhu2022celebvhq} &   3,\,668    &  12,\,000  &      4.00       &        13    \\
CelebV-Text~\cite{yu2022celebvtext} & 9,\,109    &   75,\,307 &        6.38      &       130    \\
\bottomrule
\end{tabular}
}
\caption{\textbf{Data statistics after curation and pre-processing.}}
\label{tab:dataset}
\end{table}

\begin{table}[ht]
\centering
\resizebox{\columnwidth}{!}{
\begin{tabular}{lrrr}
\toprule
Dataset         & \# Videos & \multicolumn{2}{c}{Duration} \\ \cmidrule(lr){3-4}
                &                        & Avg. (sec.) & Total (hrs.)  \\ \midrule
HDTF~\cite{hdtf}         &   318    &    139.08      &       12     \\
CelebV-HQ~\cite{zhu2022celebvhq}    &  35,\,666  &      6.86       &        68    \\
CelebV-Text~\cite{yu2022celebvtext}   &   70,\,000 &       14.35      &       279    \\
\bottomrule
\end{tabular}
}
\caption{\textbf{Data statistics before curation.}}
\label{tab:uncleaned_dataset}
\end{table}

\section{Implementation details}

\paragraph{Code}

The code and model weights will be released upon acceptance.

\paragraph{Hyperparameters \& Training Configuration}



We summarize all the hyperparameters of our pipeline in Table~\ref{tab:model_params}. The weights of the U-Net and VAE are initialized from SVD~\cite{blattmann2023stablevideodiffusionscaling}. The interpolation model undergoes more training steps because its task differs more significantly from the original task of SVD.

\begin{table}[ht]
\centering
\resizebox{\columnwidth}{!}{
\begin{tabular}{lc}
\toprule
Hyperparameter                & Value \\ \midrule
Keyframe sequence length ($T$)       & 14       \\
Keyframe spacing ($S$)               & 12       \\
Interpolation sequence length ($S$)  & 12       \\
Keyframe training steps              & 60,\,000   \\
Interpolation training steps         & 120,\,000 \\
Training batch size                  & 32             \\
Optimizer                            & AdamW          \\
Learning rate                        & $1 \times 10^{-5}$ \\
Warmup steps                        & 1,\,000          \\
Inference steps                      & 10       \\
GPU used                             & NVIDIA A100    \\
Video frame rate & 25 \\
Audio sample rate & 16,\,000 \\
Resolution & $512 \times 512$  \\
Pixel loss weighting ($\lambda_2$) & 1 \\
Audio condition drop rate for CFG~\cite{Ho2022ClassifierFreeDG}            & 20\,\%                \\
Identity condition drop rate for CFG~\cite{Ho2022ClassifierFreeDG}        & 10\,\%                 \\
\bottomrule
\end{tabular}}
\caption{\textbf{Default model hyperparameters and training configurations.}}
\label{tab:model_params}
\end{table}

\section{LipLeak}

\begin{figure}
    \centering
    \includegraphics[width=\linewidth]{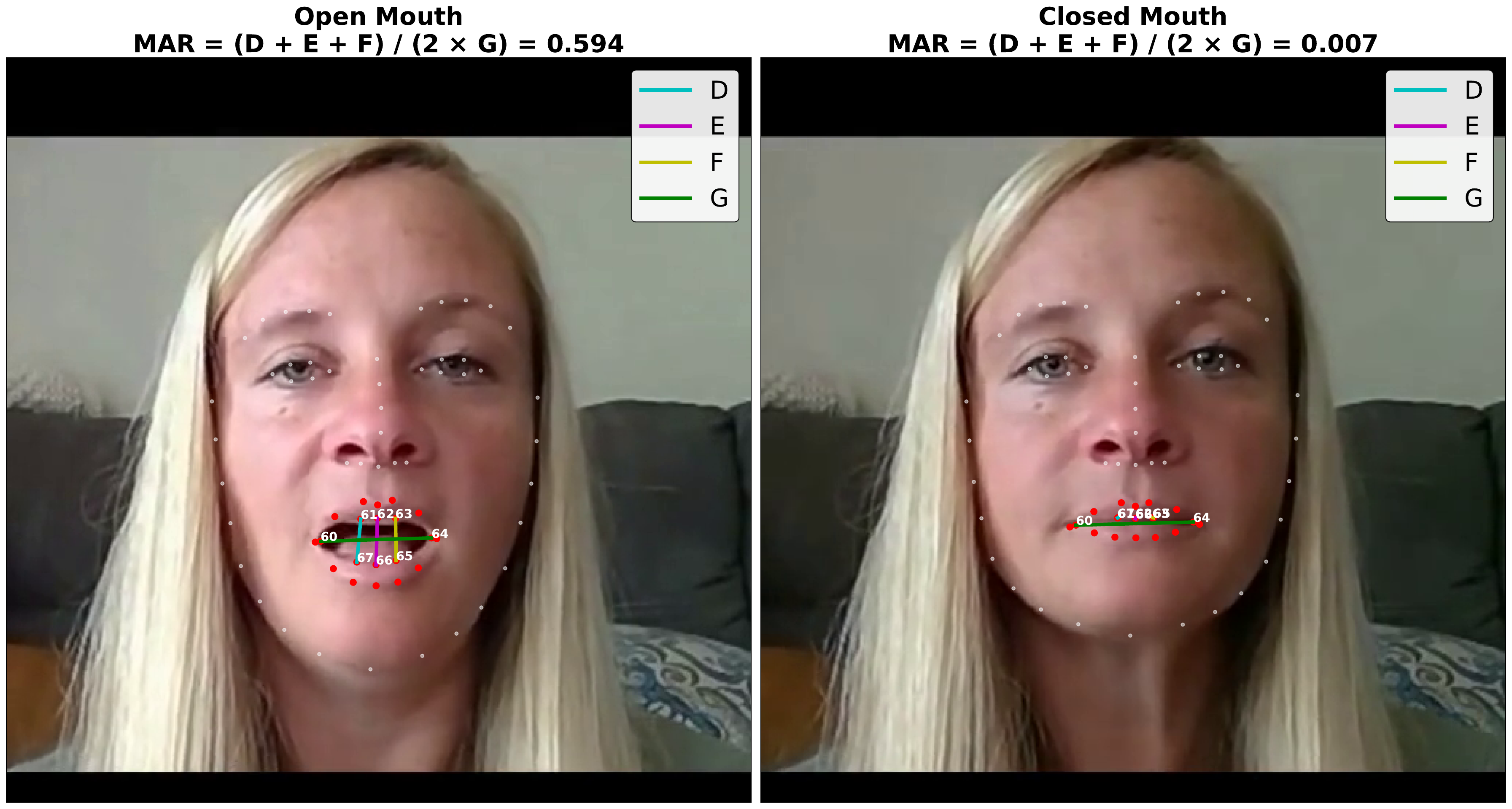}
    \caption{\textbf{LipLeak measurement example.}}
    \label{fig:lipleak_supp}
\end{figure}

We introduce LipLeak as part of our evaluation pipeline for measuring expression leakage. The first step in computing LipLeak is to calculate the mouth aspect ratio (MAR) from facial landmarks, as illustrated in Figure~\ref{fig:lipleak_supp}. This ratio quantifies the vertical openness of the mouth relative to its width, increasing as the mouth opens wider. Since LipLeak is based on a ratio, it's a a scale-invariant measure, and allows for consistent evaluation across different video resolutions and face sizes. To determine whether the mouth is open or closed, we define a threshold for the MAR. In our case, we select a threshold of 0.25, as any MAR below this value consistently represents a closed mouth based on visual inspection of several samples.

\begin{figure}
    \centering
    \includegraphics[width=\linewidth]{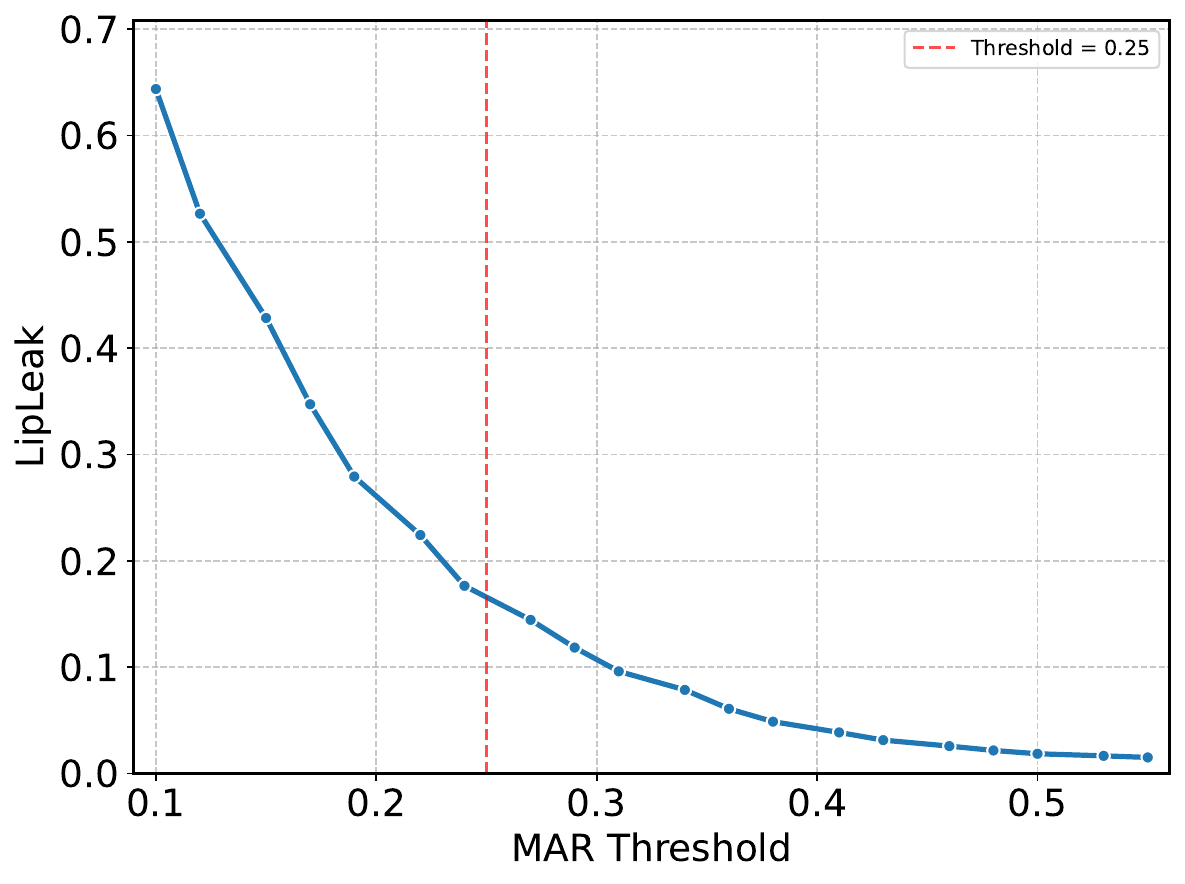}
    \caption{\textbf{LipLeak as a function of the MAR threshold.}}
    \label{fig:lipleak_thresh}
\end{figure}

To assess the sensitivity of LipLeak to threshold selection, we analyze how the metric behaves when the threshold is varied linearly (Figure~\ref{fig:lipleak_thresh}). The results show a continuous decrease in LipLeak as the threshold increases. This predictable behavior is essential, as it ensures that LipLeak can serve as a reliable and interpretable metric for evaluating mouth leakage across different conditions. A well-behaved metric should exhibit smooth variations with respect to its parameters, preventing erratic jumps or inconsistencies that could compromise its usability in quantitative evaluations.

\section{Occlusion handling}

\begin{figure}
    \centering
    \includegraphics[width=\linewidth]{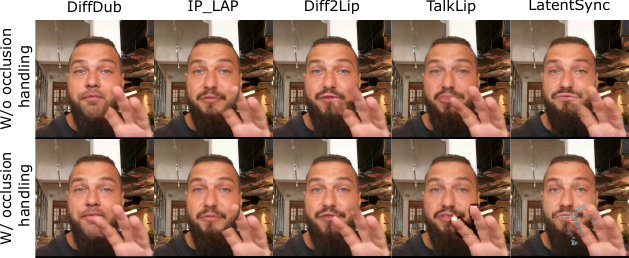}
    \caption{\textbf{Effectiveness of Occlusion Handling Across Different Methods.}}
    \label{fig:occlusions_others}
\end{figure}

We propose a method to handle occlusions at inference time, eliminating the need for model retraining to apply our technique. This makes our approach highly flexible and adaptable across different methods. Figure~\ref{fig:occlusions_others} illustrates the application of our occlusion handling technique to several existing methods:
\begin{itemize}
    \item \textbf{DiffDub~\cite{DiffDub} and Diff2Lip~\cite{diff2lip}:} Our approach works out of the box, seamlessly handling occlusions without requiring modifications.
    \item \textbf{LatentSync~\cite{latentsync}:} Since this method employs a fixed mask, the model has never been exposed to variations in masking. As a result, it struggles to adapt to the new mask patterns introduced by our occlusion-handling technique, highlighting a key drawback of using a rigid masking approach.
    \item \textbf{IP\_\,LAP~\cite{Zhong_2023_CVPR}:} This model generates the mouth region separately through an audio-to-landmark module. Consequently, the occlusion mask has no direct effect, and the mouth is generated on top of the occlusion.
    \item \textbf{TalkLip~\cite{talklip}:} At first glance, TalkLip appears to function without occlusion handling. However, it achieves this by concatenating frames from the original video to generate new frames. This shortcut enables occlusion handling but comes at the cost of significant expression leakage, as evidenced by its very high LipLeak score in Table~\ref{tab:metrics_comparison}.
\end{itemize}

\section{User study results}

\begin{figure}
\centering
\includegraphics[width=\linewidth]{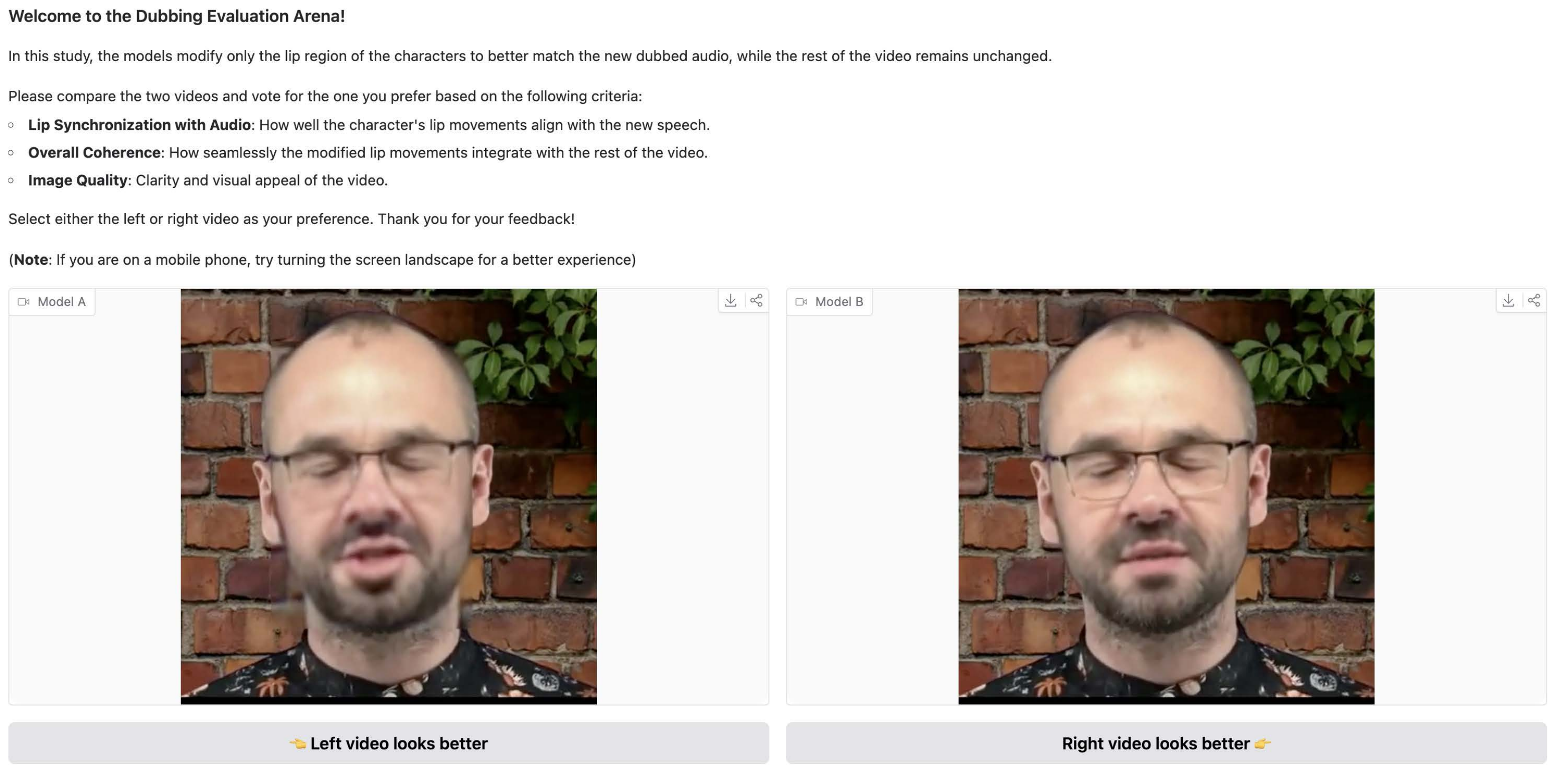}
\caption{\textbf{User study interface.} Participants were shown side-by-side videos and asked to select the preferred one based on lip synchronization, coherence, and quality.}
\label{fig:scree_user}
\end{figure}

To ensure that the objective metrics presented in Table~\ref{tab:metrics_comparison} align with human perception, we conduct a user study to evaluate model performance in terms of lip synchronization, overall coherence, and image quality. Participants are presented with pairs of videos and asked to select the one they preferred based on these criteria. The video pairs are randomly sampled from the pool of models listed in Table~\ref{tab:metrics_comparison} to ensure a fair and unbiased comparison. A total of 1,\,000 pairwise comparisons were collected, providing a robust dataset for evaluating human preferences. Figure~\ref{fig:scree_user} shows a screenshot of the user study interface, illustrating the evaluation setup.

\paragraph{Elo ratings}

To assess the relative performance of different models in our evaluation framework, we employ the Elo rating system~\cite{elo1978rating}, a widely used method for ranking competitors based on pairwise comparisons. The Elo rating system assigns scores to models based on their performance in direct comparisons, updating their ratings dynamically as more results are collected.

\begin{figure}
\centering
\includegraphics[width=\linewidth]{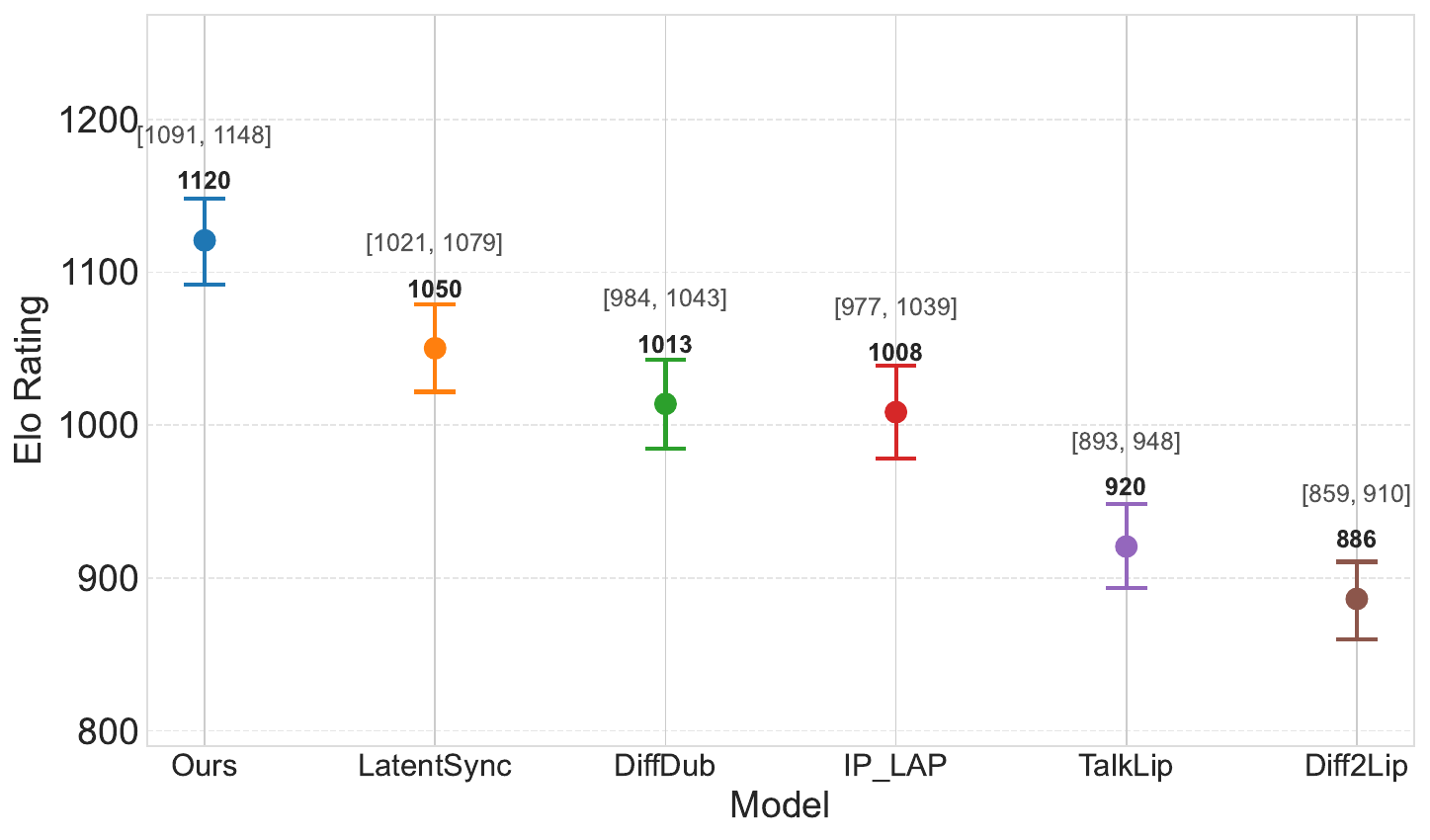}
\caption{\textbf{Elo ratings in the reconstruction setting.} Higher ratings indicate better performance in generating videos with original audio as input.}
\label{fig:elo_recons}
\end{figure}

\begin{figure}
\centering
\includegraphics[width=\linewidth]{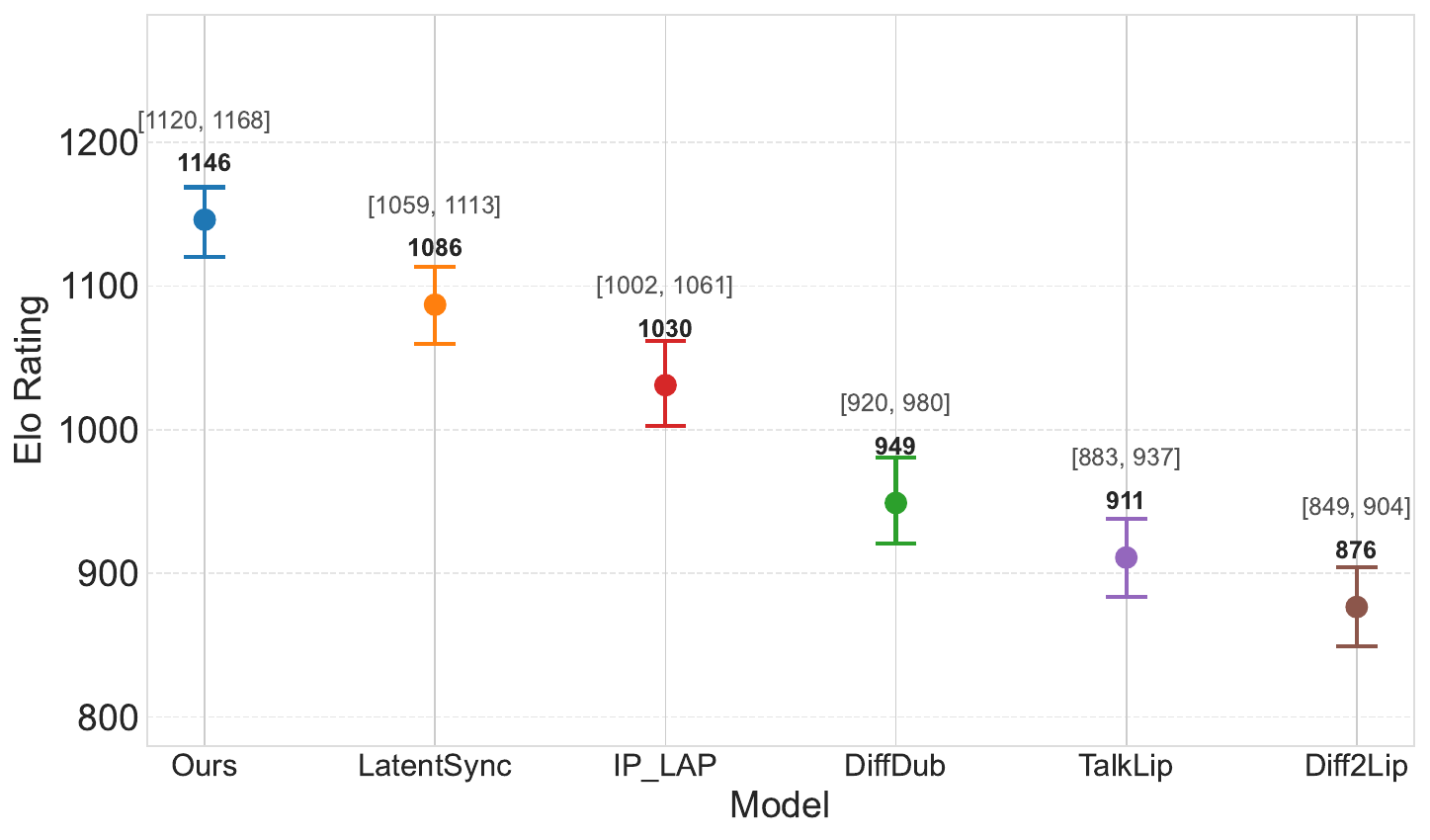}
\caption{\textbf{Elo ratings in the cross-sync setting.} Higher ratings indicate better performance in generating videos with different audio from input.}
\label{fig:elo_cross}
\end{figure}

We evaluate Elo ratings in two distinct settings:
\begin{itemize}
    \item \textbf{Reconstruction setting (Figure~\ref{fig:elo_recons}):} In this scenario, we compare videos are generated using the same audio as in the original video.
    \item \textbf{Cross-Synchronization Setting (Figure~\ref{fig:elo_cross}):} In this scenario, we compare videos generated using a different audio from the original video.
\end{itemize}

In both cases, our model consistently outperforms competing methods, achieving higher Elo ratings. This demonstrates its superior ability to generate high-quality, accurately synchronised lip movements, both in the reconstruction and cross-synchronization tasks.

\paragraph{Elo rating distributions}

\begin{figure}
\centering
\includegraphics[width=\linewidth]{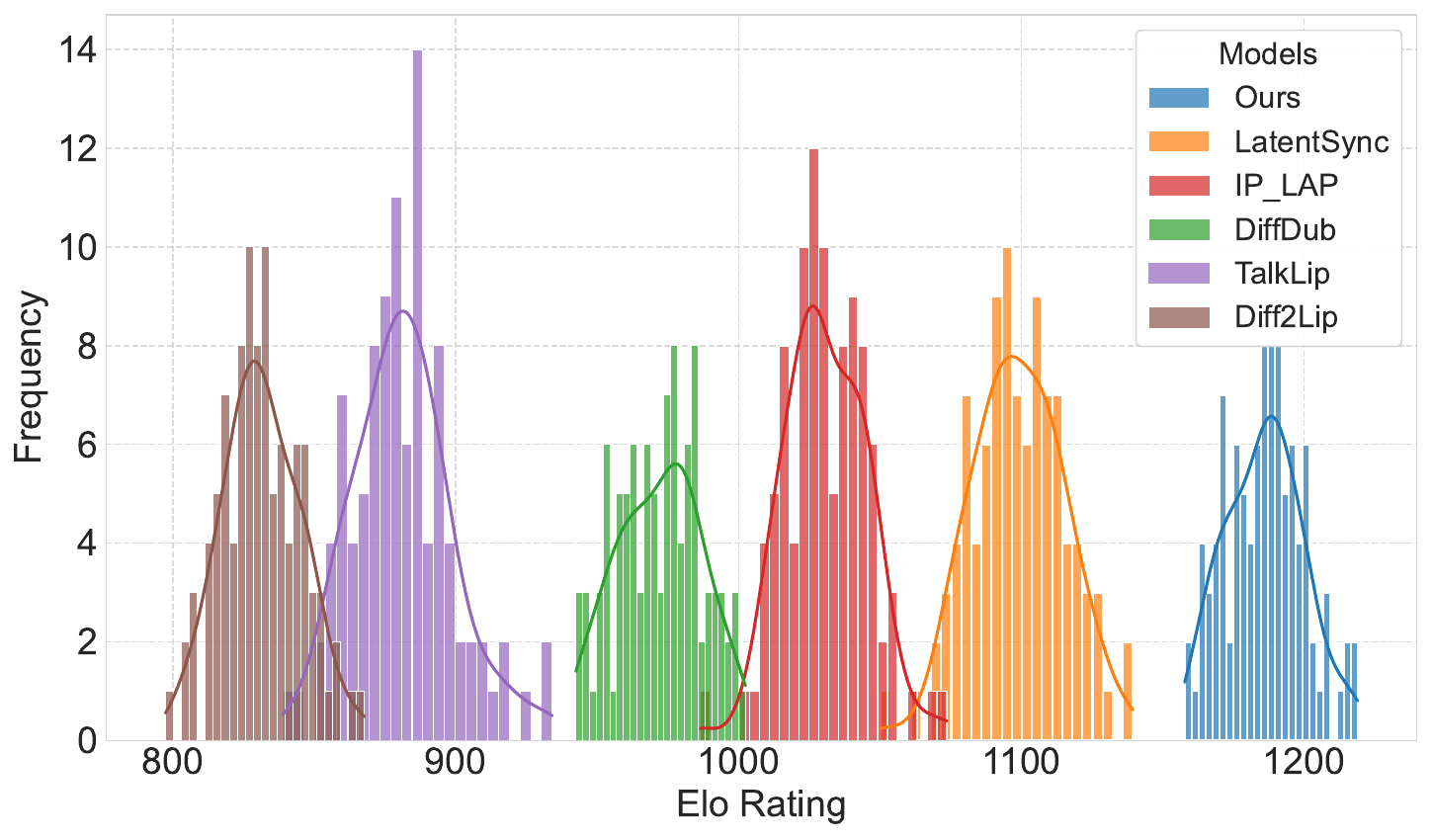}
\caption{\textbf{Distribution of Elo ratings across all evaluated models.} This histogram illustrates the spread of Elo scores, highlighting performance gaps or clustering amongst different models.}
\label{fig:elo_hist}
\end{figure}

To better understand the distribution and variance of model rankings, we analyse the overall Elo ratings across all evaluated models. Figure~\ref{fig:elo_hist} presents a histogram of Elo scores, illustrating how models are ranked relative to each other. A well-separated distribution suggests clear performance differences between models, whereas overlapping scores indicate models with similar performance levels. Our model achieves the highest Elo ratings, forming a well-defined peak that highlights its superior performance. In contrast, baseline models display varying degrees of separation, with some exhibiting significant overlap, suggesting closer competition and comparable performance in certain cases.

\paragraph{Win rates}

\begin{figure}
\centering
\includegraphics[width=\linewidth]{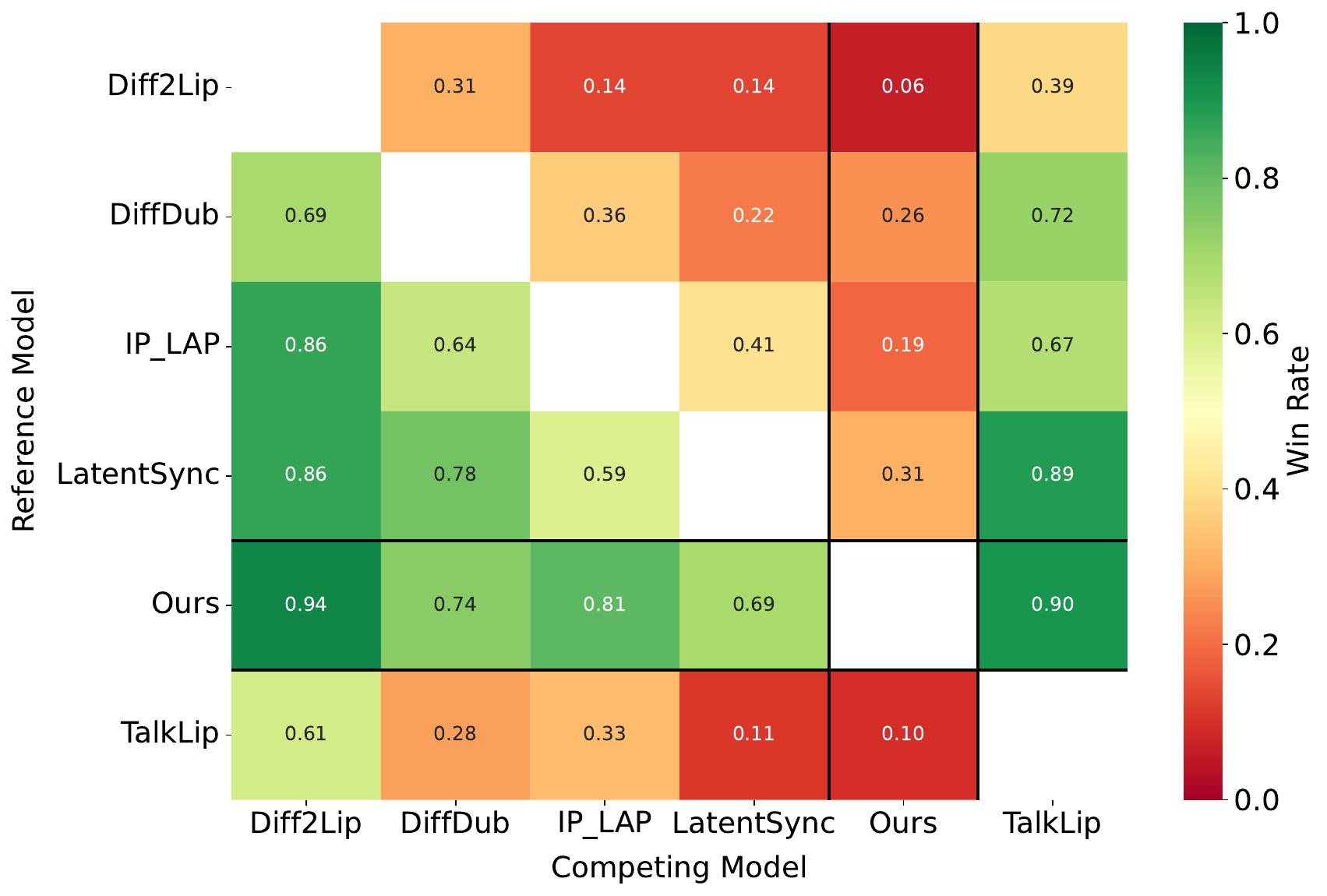}
\caption{\textbf{Win rate matrix for pairwise model comparisons.} Each cell represents the proportion of matchups where one model outperforms another, offering insight into head-to-head performance.}
\label{fig:elo_win}
\end{figure}

Beyond Elo ratings, we compute win rates to assess how often each model outperforms others in pairwise comparisons. The win rate matrix in Figure~\ref{fig:elo_win} provides a detailed overview of direct matchups, where each cell represents the percentage of times one model wins against another. This analysis helps identify dominant models and potential inconsistencies in ranking. Our model consistently outperforms competing approaches, achieving a minimum win rate of 69\,\% and a maximum of 94\,\%. These results indicate a strong and reliable performance advantage over alternative methods.

\section{Additional ablations}



\paragraph{Guidance}

Guidance plays a crucial role in the performance of diffusion models~\cite{dhariwal2021diffusionmodelsbeatgans, ho2020ddpm}. In our case, we use a modified version of Classifier-Free Guidance (CFG)~\cite{Ho2022ClassifierFreeDG}, which applies separate scaling factors to the audio and identity conditions. Specifically, our guidance function is defined as follows:
\begin{equation} \label{eq:D_function}
        z = z_{\emptyset} + w_{\text{id}} \cdot (z_{\text{id}} - z_{\emptyset}) + w_{\text{aud}} \cdot (z_{\text{id \& aud}} - z_{\text{id}}),
\end{equation}
where:
\begin{itemize}
    \item $w_{\text{aud}}$ and $w_{\text{id}}$ are the guidance scales for audio and identity, respectively.
    \item $z_{\emptyset}$ represents the model output when all conditions are set to 0.
    \item $z_{\text{id}}$ is the output when only the identity condition is applied.
    \item $z_{\text{id \& aud}}$ is the output when both audio and identity conditions are applied.
\end{itemize}

By separating the audio and identity guidance conditions, we enable more control over the generated videos, ultimately leading to improved performance. Experimentally, we found that setting $w_{\text{aud}}=5$ and $w_{\text{id}}=2$ yields the best results. This configuration achieves a 29.73\,\% improvement in LipScore, significantly enhancing lip synchronization accuracy. While this comes at a 14.75\,\% increase in CMMD and a minor 2.80\,\% increase in FVD, the overall perceptual quality remains strong, making this trade-off highly beneficial for generating realistic and synchronized videos. We summarize these results in Table~\ref{tab:guidance_table}, demonstrating the effectiveness of our approach compared to standard CFG.

\begin{table}[ht]
\centering
\resizebox{\columnwidth}{!}{%
\begin{tabular}{lcccc}
\toprule
Guidance     &        CMMD $\downarrow$& FVD $\downarrow$& LipScore $\uparrow$  \\ \midrule
CFG      &  \textbf{0.061}  &  \textbf{200.71}   & 0.37  \\
\rowcolor{Gray!40} Ours ($w_{\text{aud}}=5$, $w_{\text{id}}=2$)   & 0.070 & 206.32 & \textbf{0.48}  \\
\bottomrule
\end{tabular}}
\caption{\textbf{Guidance ablation} in the cross-sync setting.}
\label{tab:guidance_table}
\end{table}

\paragraph{Losses}

\begin{figure}
\centering
\includegraphics[width=\linewidth]{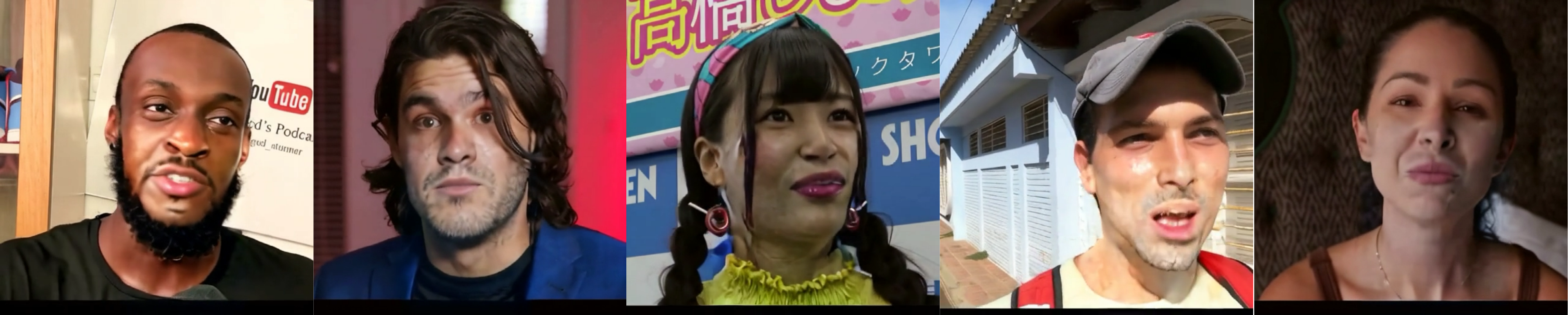}
\caption{\textbf{Examples of inconsistent mouth regions obtained by training with an additional LPIPS pixel loss.}}
\label{fig:lpips}
\end{figure}

We present an ablation on the impact of applying a pixel loss in addition to the diffusion loss in Table~\ref{tab:pixel_table}. Our findings indicate that adding a $L_2$ loss in pixel space leads to a slight improvement in image and video quality while maintaining the same level of lip synchronization. However, contrary to the findings in \cite{keyface}, we did not find that adding an additional LPIPS pixel loss benefits the model. Instead, it causes the mouth region to deviate too much from the rest of the image, as illustrated in Figure~\ref{fig:lpips}. This discrepancy arises because facial animation is a different task from lip synchronization, with the latter being more closely related to an inpainting task rather than full facial reconstruction.

\begin{table}[ht]
\centering
\begin{tabular}{lcccc}
\toprule
Loss     &        CMMD $\downarrow$& FVD $\downarrow$& LipScore $\uparrow$  \\ \midrule
No pixel loss      &  0.075  &  215.71   & \textbf{0.48}  \\
\rowcolor{Gray!40} $L_2$  & \textbf{0.070} & \textbf{206.32} & \textbf{0.48}  \\
\bottomrule
\end{tabular}
\caption{\textbf{Pixel loss ablation} in the cross-sync setting.}
\label{tab:pixel_table}
\end{table}

\section{Limitations}

\begin{figure}
\centering
\includegraphics[width=\linewidth]{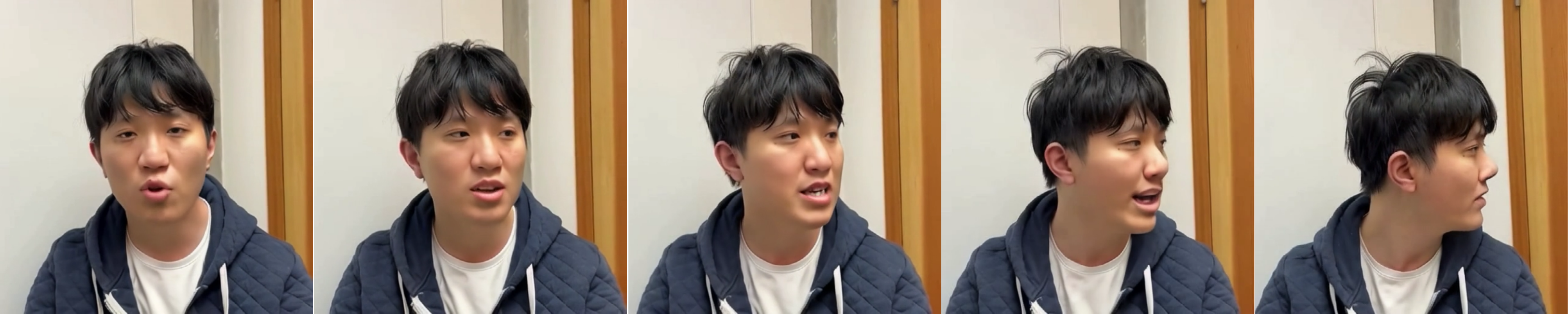}
\caption{\textbf{Examples of generated videos at different angles.}}
\label{fig:example_angles}
\end{figure}

To assess the limitations of our approach, we construct a small dataset consisting of seven identities, where each individual recites the same two sentences at five different angles: 0°, 20°, 45°, 70°, and 90°, as illustrated in Figure~\ref{fig:example_angles}. This setup allows us to systematically evaluate how the model performs under varying viewpoint conditions.

\begin{figure}
\centering
\includegraphics[width=\linewidth]{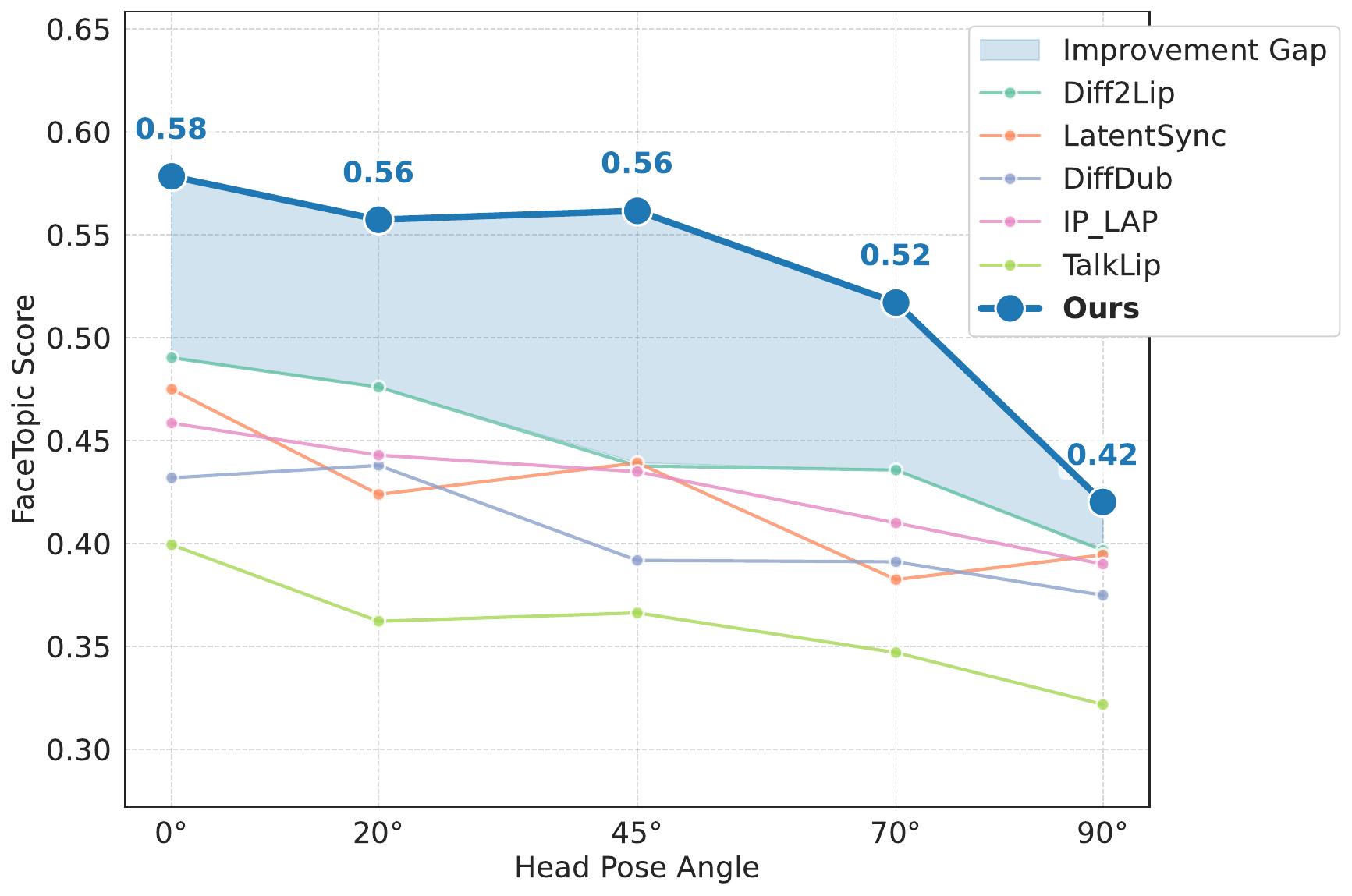}
\caption{\textbf{Impact of head pose on model performance.}}
\label{fig:results_edge}
\end{figure}

We present the results of TOPIQ~\cite{topiq} with respect to the angle in Figure~\ref{fig:results_edge}. We use TOPIQ because it is a no-reference image quality metric that does not require a large ground-truth dataset for direct comparison, making it more practical than FID or FVD, which rely on reference distributions that may be skewed or incomplete across extreme angles. Additionally, unlike variance of Laplacian (VL), which only captures blurriness, TOPIQ provides a more comprehensive measure of perceptual quality degradation, including semantic distortions that become more pronounced at oblique head poses. The results indicate that all approaches exhibit performance degradation as the angle increases. This is a key limitation of our model, which is also observed across baseline methods. This decline in performance can be attributed to the inherent biases in our training datasets, which predominantly contain frontal faces. As a result, the model struggles to infer occluded or unseen facial regions when presented with extreme head poses. One potential solution is to provide identity frames from multiple viewpoints during training, allowing the model to learn a more comprehensive facial representation. However, this would require extesnsive new data collection and further investigation, and is therefore left for future work.

\section{Additional qualitative results}

\begin{figure*}
    \centering
    \includegraphics[width=\linewidth]{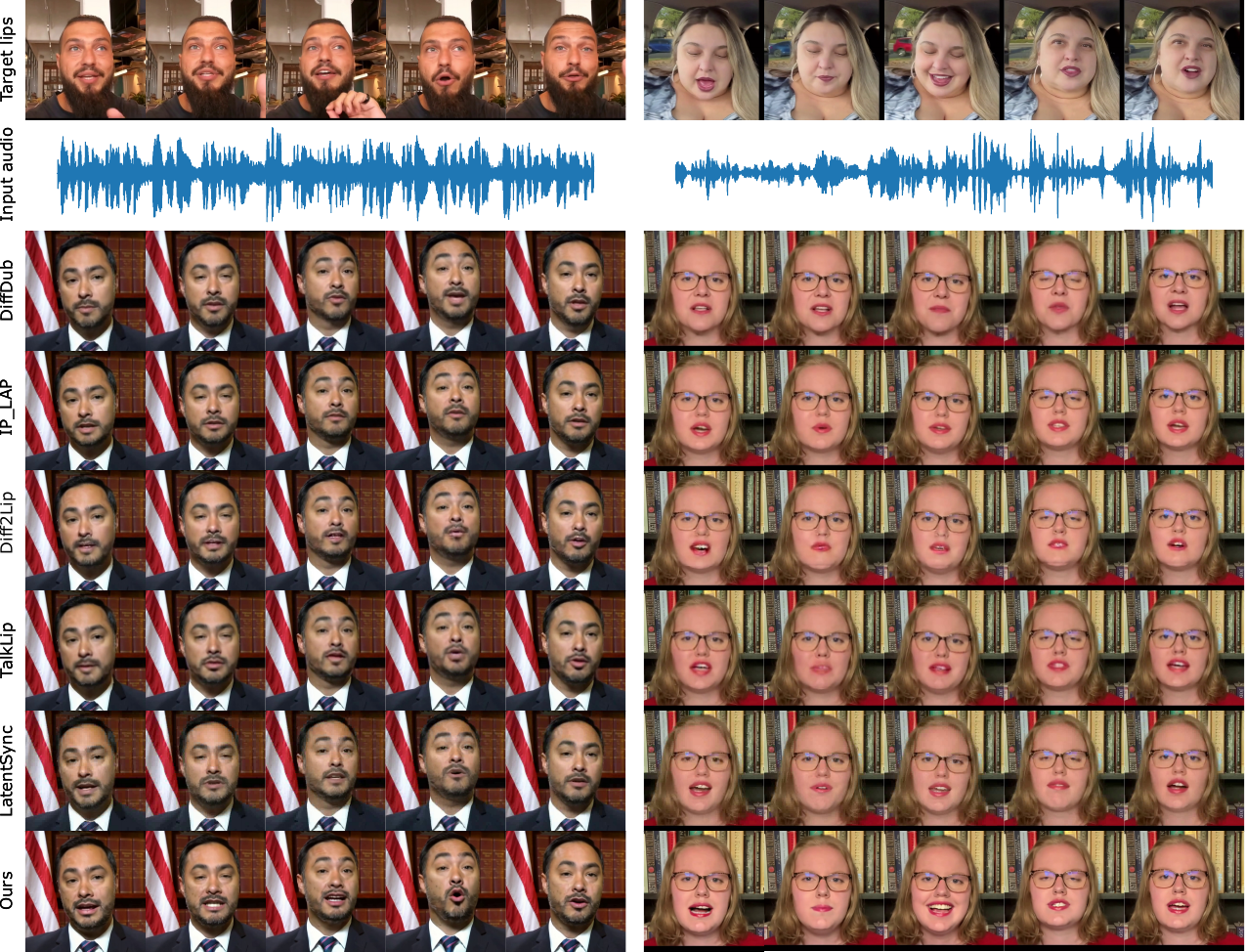}
    \caption{\textbf{Additional qualitative comparison.}}
    \label{fig:qualitative_supp_extra}
\end{figure*}

\begin{figure*}
    \centering
    \includegraphics[width=\linewidth]{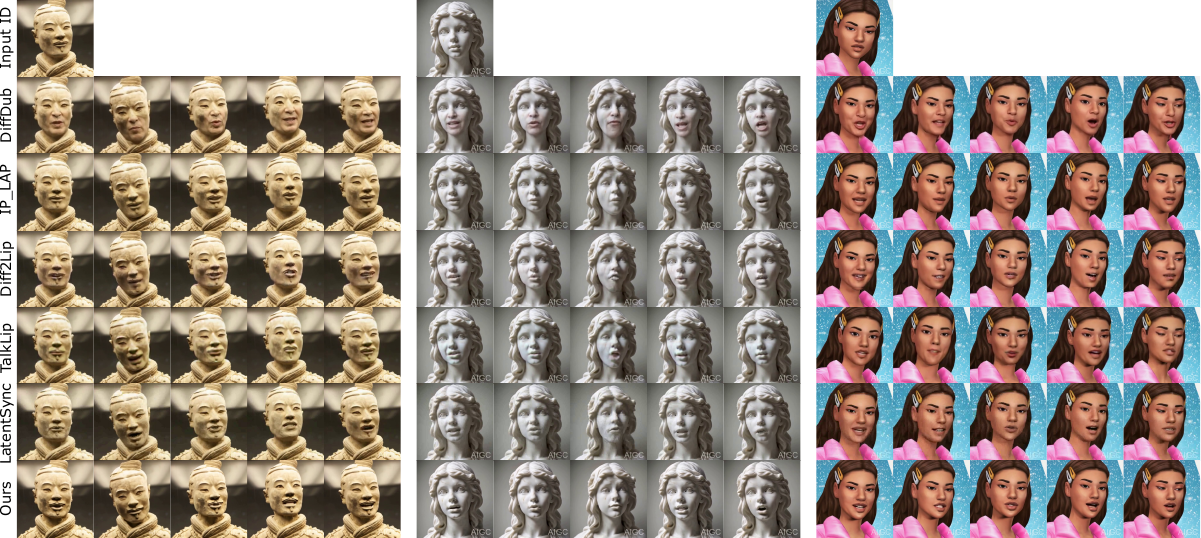}
    \caption{\textbf{Qualitative comparison on non-human ids.}}
    \label{fig:qualitative_non_human}
\end{figure*}

We present additional qualitative results in Figure~\ref{fig:qualitative_supp_extra}. As reported in the main paper, our model demonstrates better alignment with the target lips while also achieving higher image quality compared to other methods. Additionally, we evaluate our model’s ability to handle non-human faces in Figure~\ref{fig:qualitative_non_human}. We find that KeySync produces plausible lip-synced animations, while competing models fail to accurately reconstruct mouth details, particularly in the first two identities, as they deviate significantly from typical human facial structures. This highlights our model’s superior adaptability in handling out-of-distribution (OOD) scenarios.

To better assess the effectiveness of our approach, we provide a series of videos as part of the supplementary material. These videos are categorized as follows:
\begin{itemize}
\item \textbf{Side-by-side comparisons:} Showcasing our method against other approaches in both reconstruction and cross-sync settings.
\item \textbf{Silent videos:} Highlighting expression leakage within the same video, demonstrating how different models handle silent audio.
\item \textbf{Occlusion cases:} Also included in the same video, presenting situations where parts of the face are obstructed, illustrating the robustness of our approach.
\item \textbf{Multilingual examples:} Evaluating the model’s performance across different languages to assess generalization.
\item \textbf{Out-of-distribution examples:} Testing our model on non-human identities, demonstrating its adaptability to non-human faces.
\item  \textbf{Examples at different angles:} Analyzing the model’s performance under varying head poses, highlighting its ability to handle different viewpoints as well as its limitations.
\item \textbf{Additional cross-sync videos:} Providing a more extensive evaluation of our model’s cross-sync capabilities across various conditions.
\end{itemize}

These supplementary videos offer a comprehensive visual demonstration of our method’s performance across a wide range of conditions.

\section{Ethical Considerations and Social Impact}

\paragraph{User study}
Our study includes a user evaluation where participants compare video outputs for lip synchronization, image quality, and coherence. All participants provided informed consent, and their responses were collected anonymously. No personally identifiable information or sensitive data were gathered, ensuring compliance with ethical research guidelines.

\paragraph{Model}
Lip-sync generation has numerous beneficial applications, including enhanced video dubbing, accessibility tools for hearing-impaired individuals, and improvements in digital content creation. However, we acknowledge that such technology can also be misused, particularly in the context of deepfake generation, which poses risks related to misinformation, identity fraud, and unethical content manipulation. To mitigate potential misuse, we emphasize that our approach is developed with a focus on fair use cases and is intended strictly for research purposes. 

\paragraph{Datasets}
We rely on publicly available datasets that were originally collected and published by external researchers. We adhere to the terms and ethical guidelines set by the dataset creators. 

\end{document}